# Modeling meaning: computational interpreting and understanding of natural language fragments

*Michael Kapustin[1], Pavlo Kapustin[2]*

[1] Moscow Institute of Physics and Technology, Department of Applied Mathematics
michael.kapustin@gmail.com

[2] University of Bergen, Department of Information and Media Studies
pkapustin@gmail.com

---

***Abstract*—In this introductory article we present the basics of an approach to implementing computational interpreting of natural language aiming to model the meanings of words and phrases. Unlike other approaches, we attempt to define the meanings of text fragments in a composable and computer interpretable way. We discuss models and ideas for detecting different types of semantic incomprehension and choosing the interpretation that makes most sense in a given context. Knowledge representation is designed for handling context-sensitive and uncertain / imprecise knowledge, and for easy accommodation of new information. It stores quantitative information capturing the essence of the concepts, because it is crucial for working with natural language understanding and reasoning. Still, the representation is general enough to allow for new knowledge to be learned, and even generated by the system. The article concludes by discussing some reasoning-related topics: possible approaches to generation of new abstract concepts, and describing situations and concepts in words (e.g. for specifying interpretation difficulties).**

## 1 Introduction

Knowledge representations based on semantic networks and frames have been criticized for imprecision and ambiguity since [1]. Formal logic-based approaches, on the other hand, do not really allow for learning new knowledge easily, as they are dependent on people codifying it [2:5-7]. Neither semantic networks nor formal logic approaches handle uncertainty or fuzzy concepts like perceptions well [3]. More importantly, mentioned representations do not contain any quantitative, computer interpretable information about the concept and relation *internal structure*, or *definitions*[1], making it virtually impossible to work with the meanings of concepts and phrases, and thus, severely restricting capabilities of the software.

We, on the other hand, tend to agree with the words of T. Winograd [4:1]: *"We assume that a computer cannot deal reasonably with language unless it can understand the subject it is discussing"*. We do not believe it is possible to create a system with understanding abilities unless we can computationally distinguish comprehension from incomprehension. We won't be able to make the system express itself meaningfully unless we have a model for the meaning itself.

The *ultimate* problem of our interest (the one we obviously do not attempt to fully solve, but consider useful to guide the design of the framework) is conducting a dialog with a system using natural language (not necessarily syntactically and grammatically correct). We understand that one needs to be more realistic than many AI researchers were in the early seventies, but, on the other hand, some steps towards modeling of the meaning were already taken (e.g. L. Zadeh, starting from [3]), and we see no other choice as to continue small steps in this direction.

---

[1] By internal structure (or definitions) here we mean something allowing computer to understand what makes this concept different (or similar) to/from other concept (and how).

This introductory article presents the basics of our approach aiming to express natural language concepts in a quantitative, computer interpretable way, and discusses some examples. Primary focus of the approach is the ability to computationally analyze the meaning of natural language statements, so that the system can analyze whether they "make sense", and interpret them. We discuss modeling different parts of speech, working with different contexts, and draft the overall phrase interpretation process, including the choice of the most sensible interpretation (when several interpretations are possible). Further, we provide some examples of criteria for assessing comprehensibility. In the last chapter, we mention some other reasoning-related aspects: first, we discuss working with abstract concepts, and then conclude with a draft of an algorithm allowing to describe situations (or concepts known to the system) in "own words". One practical use case for such description is communicating interpretation difficulties to the user.

# 2 Related work

*"The meaning of meaning and how to deal with meaning in formal and natural systems has been one of the great mysteries of intelligence - artificial or otherwise. It has been an issue from the earliest days of philosophy and logic, and it has become an engineering issue with the advent of computerized question answering systems, information retrieval systems, machine translation, speech understanding, intelligent agents, and other applications of natural language processing, knowledge representation, and artificial intelligence in general"* [5:75].

*Computational* aspect of *meaning* seems to remain a mystery for the most part of it. There is fairly little research around quantitative modeling of the meaning of natural language constructs.

In Quantitative Fuzzy Semantics [3], L. Zadeh asks: *"Can the fuzziness of meaning be treated quantitatively, at least in principle?"* He suggests modeling concepts like "young", "close to middle-age" and "middle-aged" as fuzzy sets.

[6] discusses using linguistic variables whose values are words or sentences in a natural language.

[7] suggests modeling linguistic hedges (e.g. "very", "more or less", "much", "essentially", "slightly") as *operators* that act on the fuzzy set representing the meaning of its operand (e.g. operator "very" acting on the meaning of operand "tall man"). Several operations for manipulating these fuzzy set representations are introduced, e.g. complementation, intersection, normalization, concentration, dilation, fuzzification, etc.

[8] discusses using fuzzy sets for modeling natural language quantifiers like "several", "most", "not many", "close to five", "approximately ten", etc.

*Test-score semantics* [9] and knowledge representation based on it [10] propose modeling the meaning of a proposition as a composition of meanings of words-elements. However, the semantics does not formalize the meaning enough (it is represented by a so-called *test procedure*) to allow handling it computationally: *"What is much more difficult, however, is to write a program which could construct an explanatory database and a test procedure without human assistance. This is a longer range problem whose complete solution must await the development of a substantially better understanding of natural languages and knowledge representation than we have at this juncture"* [9:33].

*Generalized Constraint Language* (GCL) [11, 12] is an evolution of test-score semantics and a basis for *Computing With Words* (CWW) [13, 14]. CWW aims to allow computations with words from natural language (instead of numbers), and it requires *precisiation* step to rewrite natural language elements in GCL. This step includes identification of *constrained variables* and *constraining relations* (and their types), together defining the meaning of the proposition [13:66-67]. This step is *"generally done by inspection"* [13:67]. In particular, no attempt is made to automatically relate the variables that are implicit in the proposition to the words in natural language. Also, same as in test-score semantics, there is no known way to programmatically generate *explanatory database*, that serves as the basis for precisiation [13:185].

In other words, CWW does not provide a way to handle the meaning computationally *before* it is precisiated. This may not be required for Computing With Words, but it is necessary for interpreting and understanding natural language, and this is what we are focusing on in this article.

# 3 Approach overview

Very briefly (and informally), the approach could be outlined like the following.

The approach is based on fuzzy logic, as it is a very good instrument for working with different levels of truthness, and concepts with unclear boundaries, phenomena commonly occurring in knowledge coming from natural language, and commonsense knowledge in particular [10].

All knowledge is encoded in *fuzzy properties* (with values ranging from zero to one), each of them encoding an independent piece of information.

*Contexts* and context hierarchies are used for structuring knowledge and modeling its context sensitivity. Context is defined as a coordinate system: If *N* is the number of independent properties in a given context, then it is said that the context contains *N axes*, and the knowledge in this context is described as a *fuzzy region* in a N-dimensional unit hypercube.

We model phrases, words and other natural language fragments as region transforms that we call *meaning-operators*. For example, a specific interpretation of a phrase is a transformation of the *source region* (the region before interpretation) by the corresponding phrase operator. The result of the transformation is what we call *resulting region* (the region after interpretation).

We see natural language understanding as choosing the interpretation that makes most sense, using different heuristics.

We can assess the meaning of an operator during its composition. We can evaluate overall phrase comprehension when the phrase operator is being applied. Then we are considering both source and resulting regions, and potentially other factors (e.g. phrase mood).

Simplified schematics of phrase interpretation is shown on Fig. 3-1.

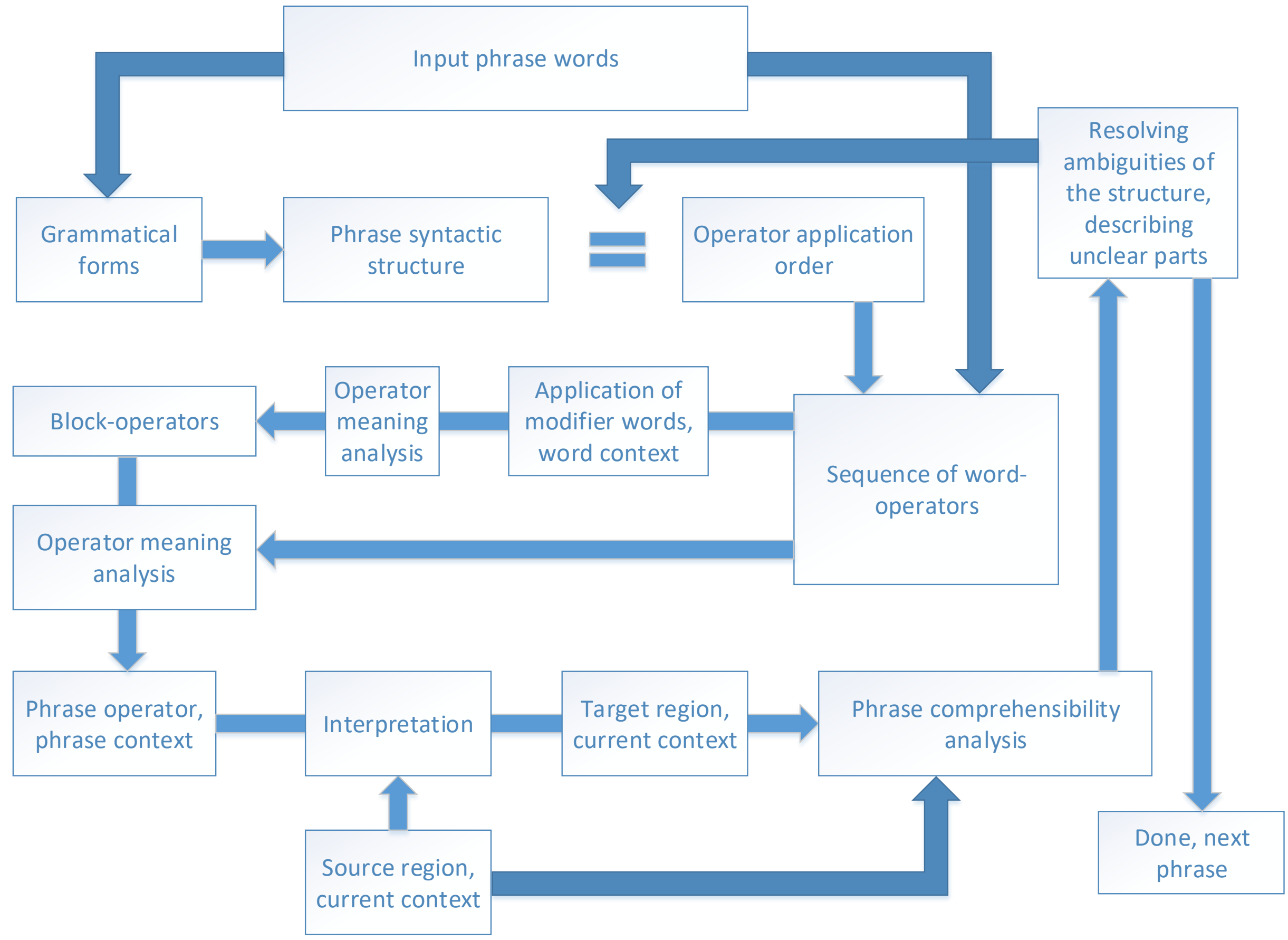

*Fig. 3-1. Simplified schematics of phrase interpretation*

# 4 Basic concepts

This chapter is an informal introduction to the concepts and ideas that we will later use to define what we mean by "meaning", and how it can be used in the phrase interpretation process.

## 4.1 Properties, context and regions

Let's assume that the state, or knowledge of a system can be completely described with a finite number of independent real parameters, or properties. Values of each property range from zero to one, so that one corresponds to maximal presence of the property, and zero corresponds to its complete absence. For example, if we would like to model vehicle speed, we could use property "quickness"[2]. "Zero" would mean "not fast at all" and "one" would mean "as fast as it gets").

We are going to call some of the properties *basic properties*. Many of the basic properties will include values that can be directly "perceived" by the system. For example, if we were developing a robot with a built-in rangefinder, the property "relative distance" would have been perceived directly. The same is true for the property "relative time", as long as the system has a built-in clock.

The rest of the properties we are going to call *derived*, with their meaning defined via other properties using the model described below.

[2] For simplicity of the examples, we are avoiding direct use of "speed", as it is a more complex concept.

*Context* is a coordinate system consisting of axes that represent values of currently relevant properties, one axis per property. For example, when talking about movement speed, two of the relevant properties could be "relative distance" and "relative time".

In the context's coordinate system we can define a fuzzy (in terms of fuzzy logic) shape: a region. Each point of the region is assigned a value between zero and one, describing this point's degree of membership.

We can use regions to express meaning of different concepts. For example, using axes *t* and *s* ("relative time" and "relative distance"), we can express regions, describing concepts "fast" and "slow" (Fig. 4-1).

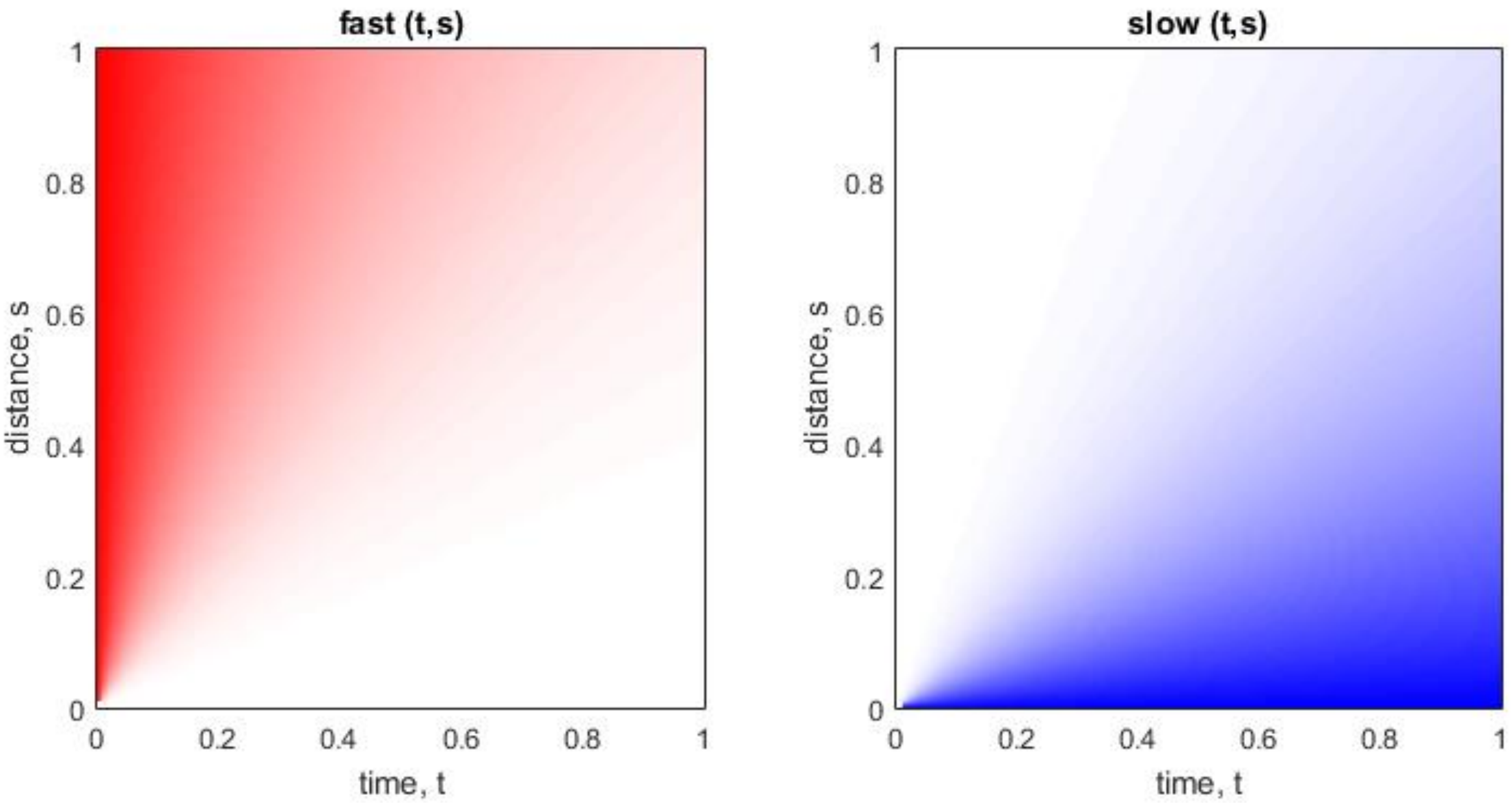

*Fig. 4-1. Concepts "fast" and "slow"*

Based on existing regions, we can introduce new (derived) properties. For example, based on the region corresponding to the concept "fast" (Fig. 4-1), we can introduce a new property "quickness".

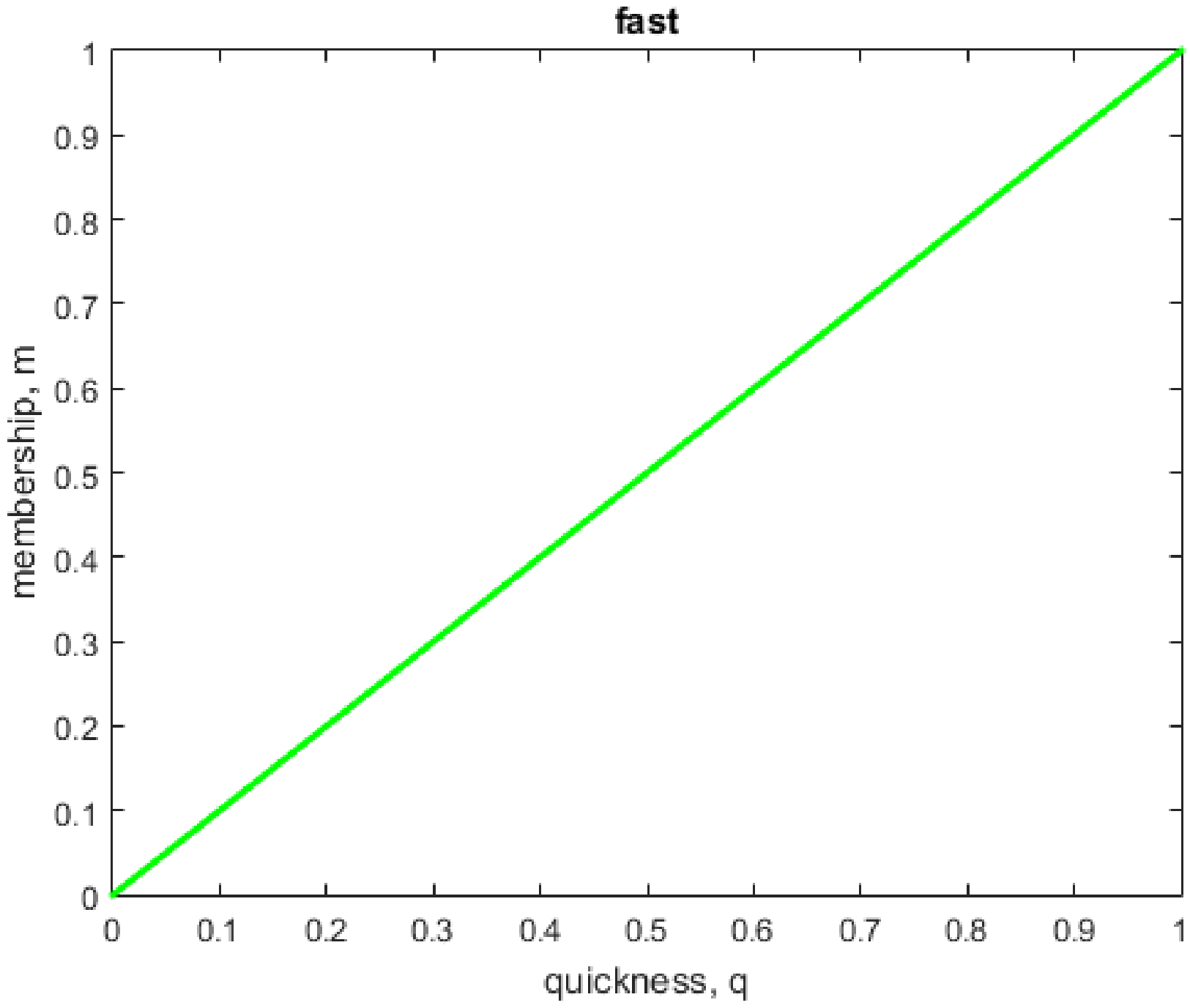

*Fig. 4-2. Concept "fast" in context with one property "quickness"*

It is natural to assume the values of "quickness" (*x*-axis on Fig. 4-2) be equal to the degrees of membership of its corresponding region's ("fast") points (color intensity on Fig. 4-1, left). As long as this relation holds, we are going to call such region *reference* region of a property.

Let's now introduce a new concept "moderatelyPaced" containing only one axis "quickness" (*q*), and define a region in this context, described by the function *moderatelyPaced(q)* (Fig. 4-3).

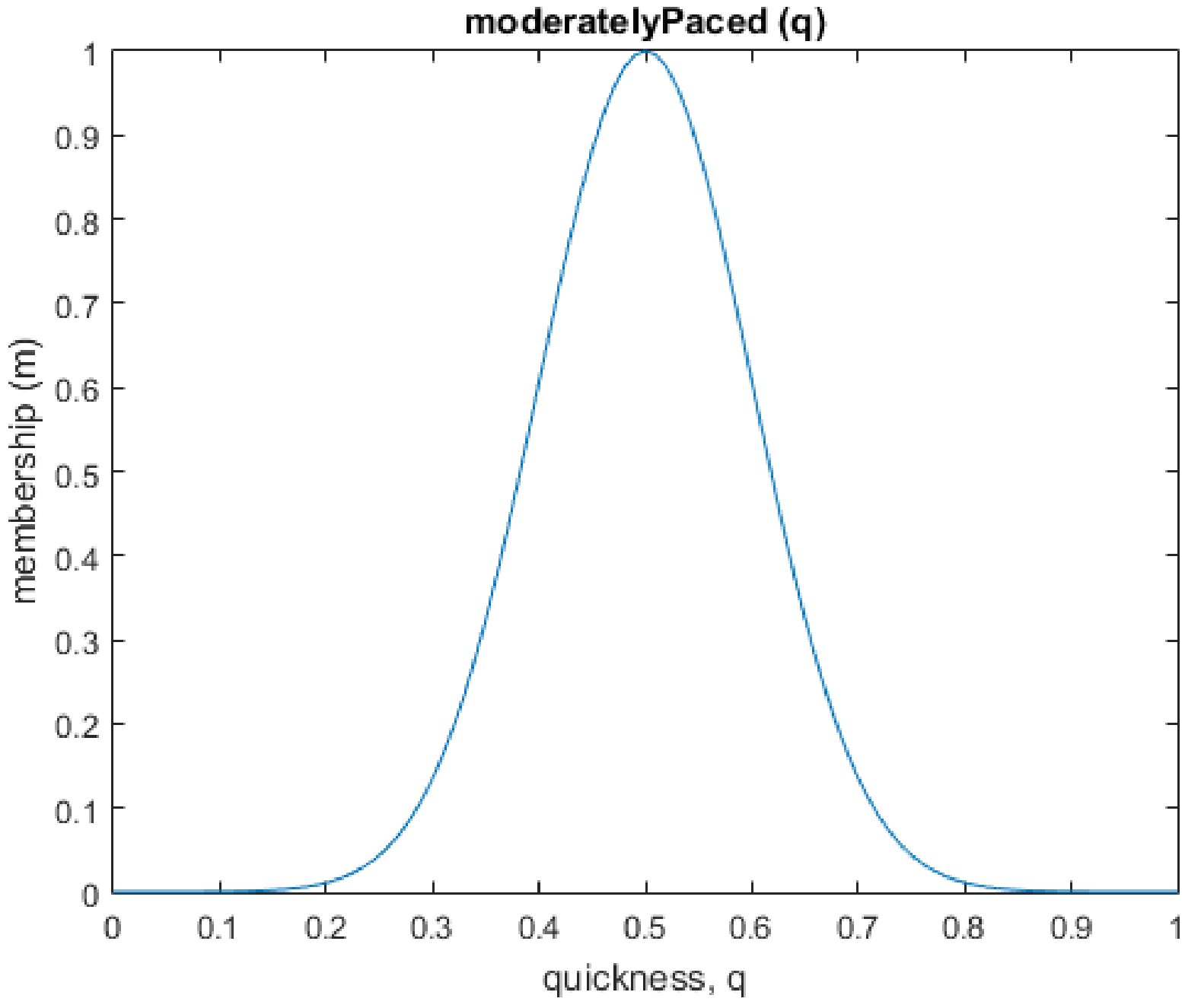

*Fig. 4-3. Concept "moderately paced"*

Please note that as long as we only have one property in this context, we are using *Y*-axis for the degree of membership (instead of using color intensity, as in the previous example).

Let's now see how this region looks in coordinates *s, t.* Remember, we assumed values of property "quickness" (*x*-axis on Fig. 4-2) to be drawn from the degrees of membership of the concept "fast" (color intensity on Fig. 4-1, left). Because of this, $moderatelyPaced(fast(s,t))$ will be function composition of $fast(s,t)$ and $moderatelyPaced(q)$, yielding a function like "$moderatelyPaced(s,t)$". This function will transform membership degree of each point of the concept $fast(s,t)$ in accordance with the rule given by $moderatelyPaced(q)$, and this corresponds to the concept "moderately paced" in the context with axes $s, t$ (Fig. 4-4).

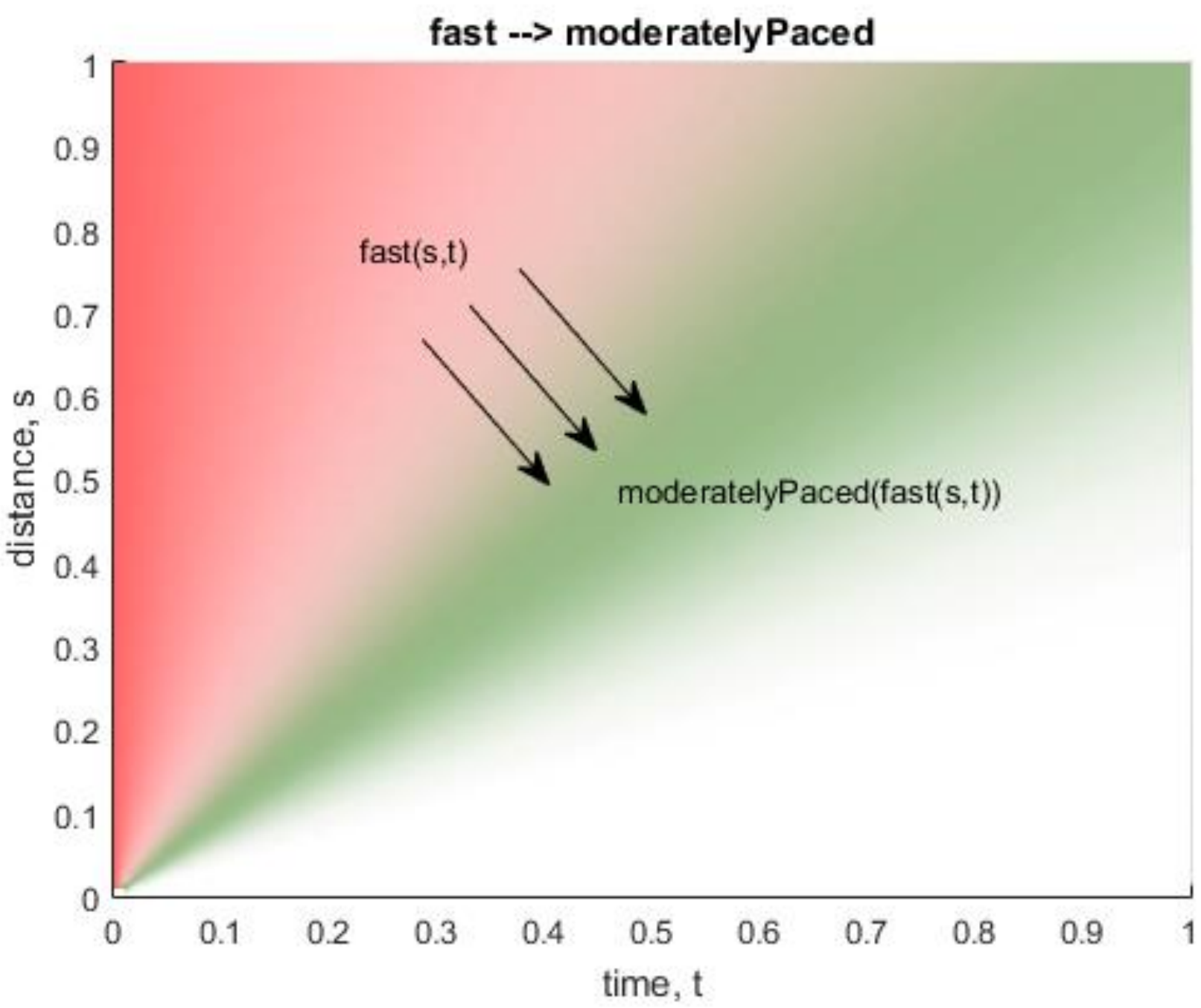

*Fig. 4-4. Mapping "moderatelyPaced" back to reference context.*

This kind of mapping is general operation, allowing us to "expand" any axis via its reference axes, thus, mapping different pieces of information into the same (reference) context for processing. We will come back to axis expansion again in 5.4.

## 4.2 Operators

Region transforms, or *operators*, are going to play key role in our model. This section is an example of two simple operators, visualizing how we can use operators for creating new concepts.

We can define the operator corresponding to the word "not" like this: $not(x) = 1 - x$, where *x* is the value of the property (Fig. 4-5). This is similar to Zadeh's "complementation" [7:10].

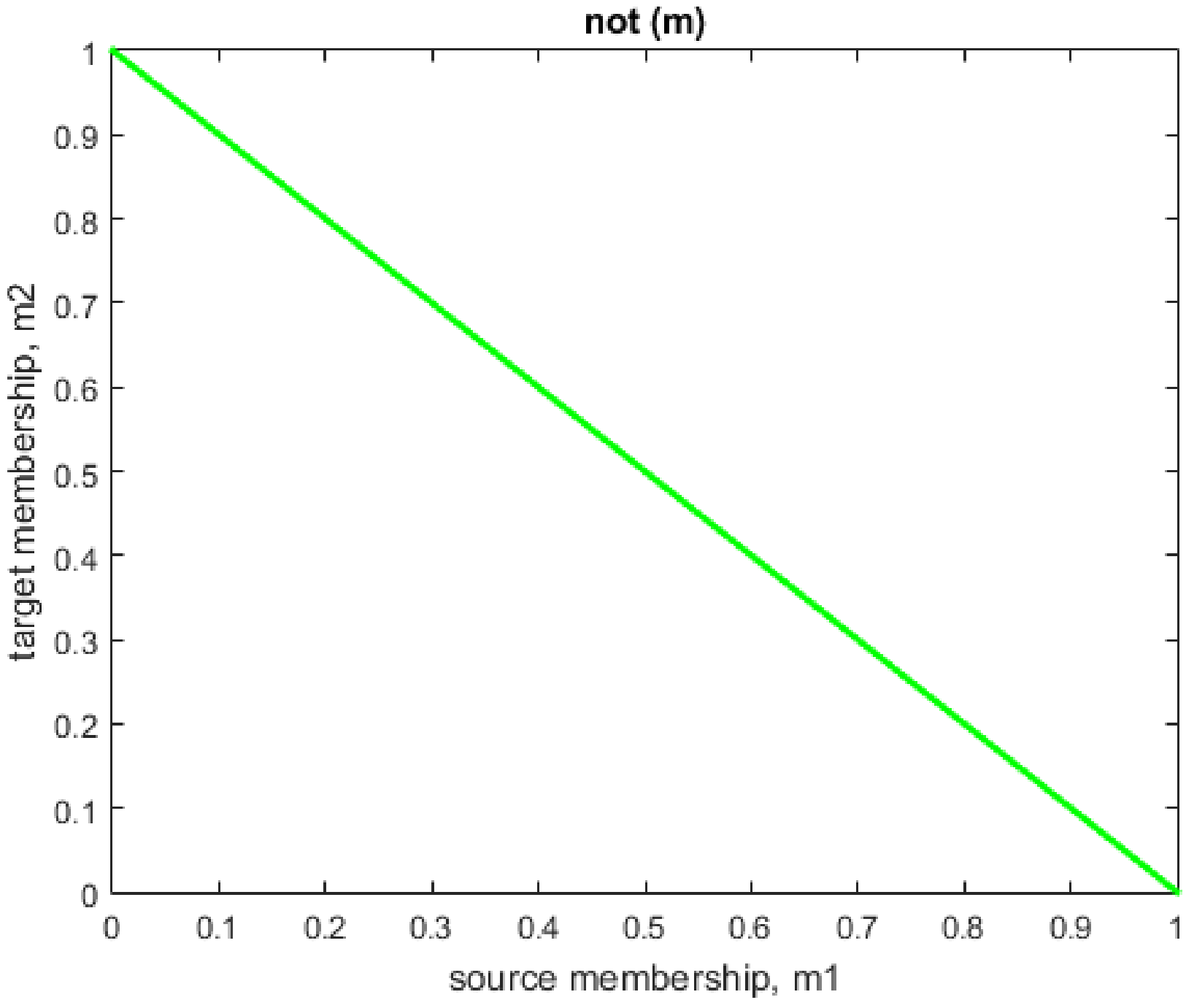

*Fig. 4-5. Operator "not"*

Using operator "not", we can define a new concept "slow" as *not(fast)* (Fig. 4-6)

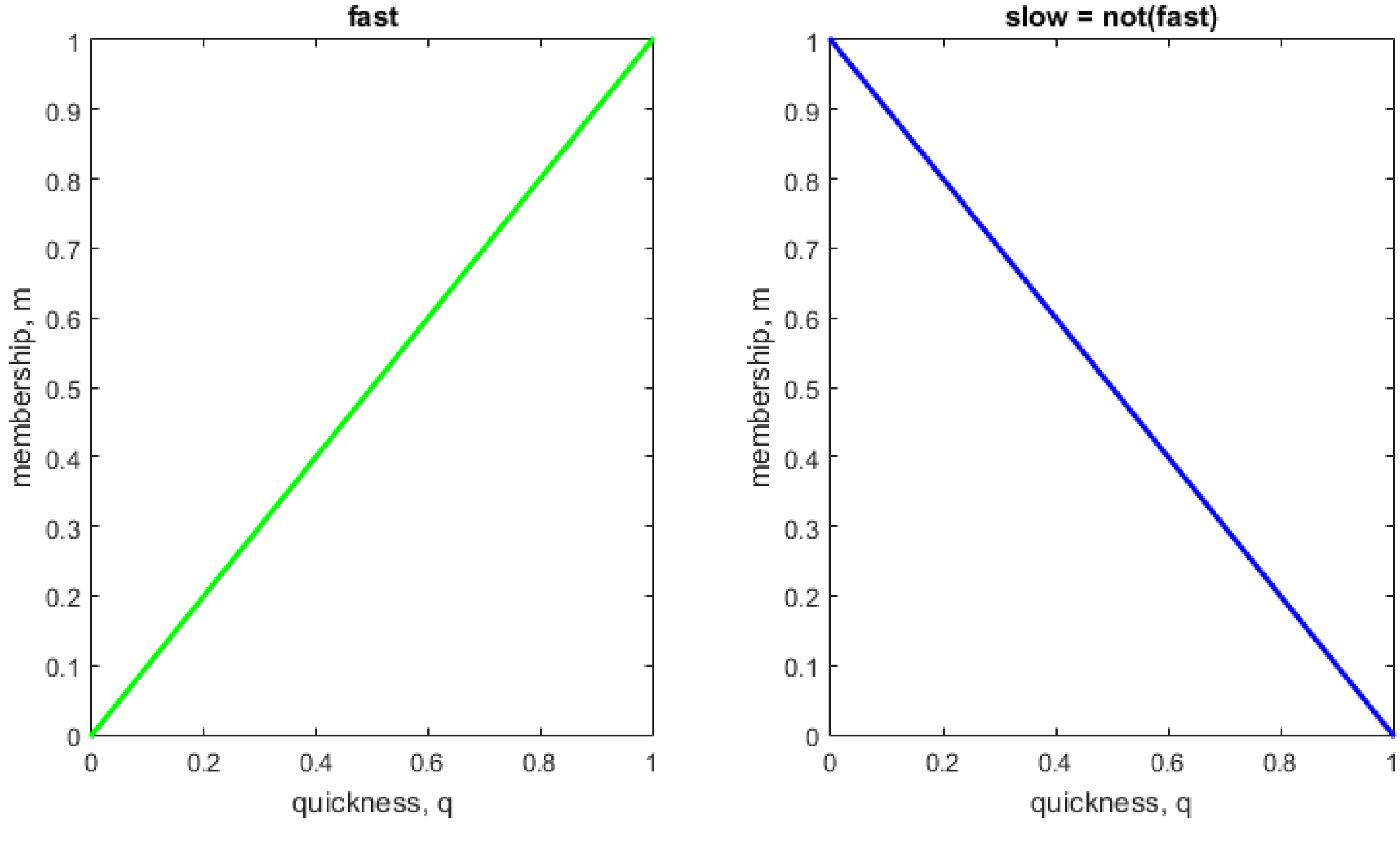

*Fig. 4-6. slow = not(fast)*

Mapping this back to our reference context, we get: $not(fast(s,t))$ (Fig. 4-7)

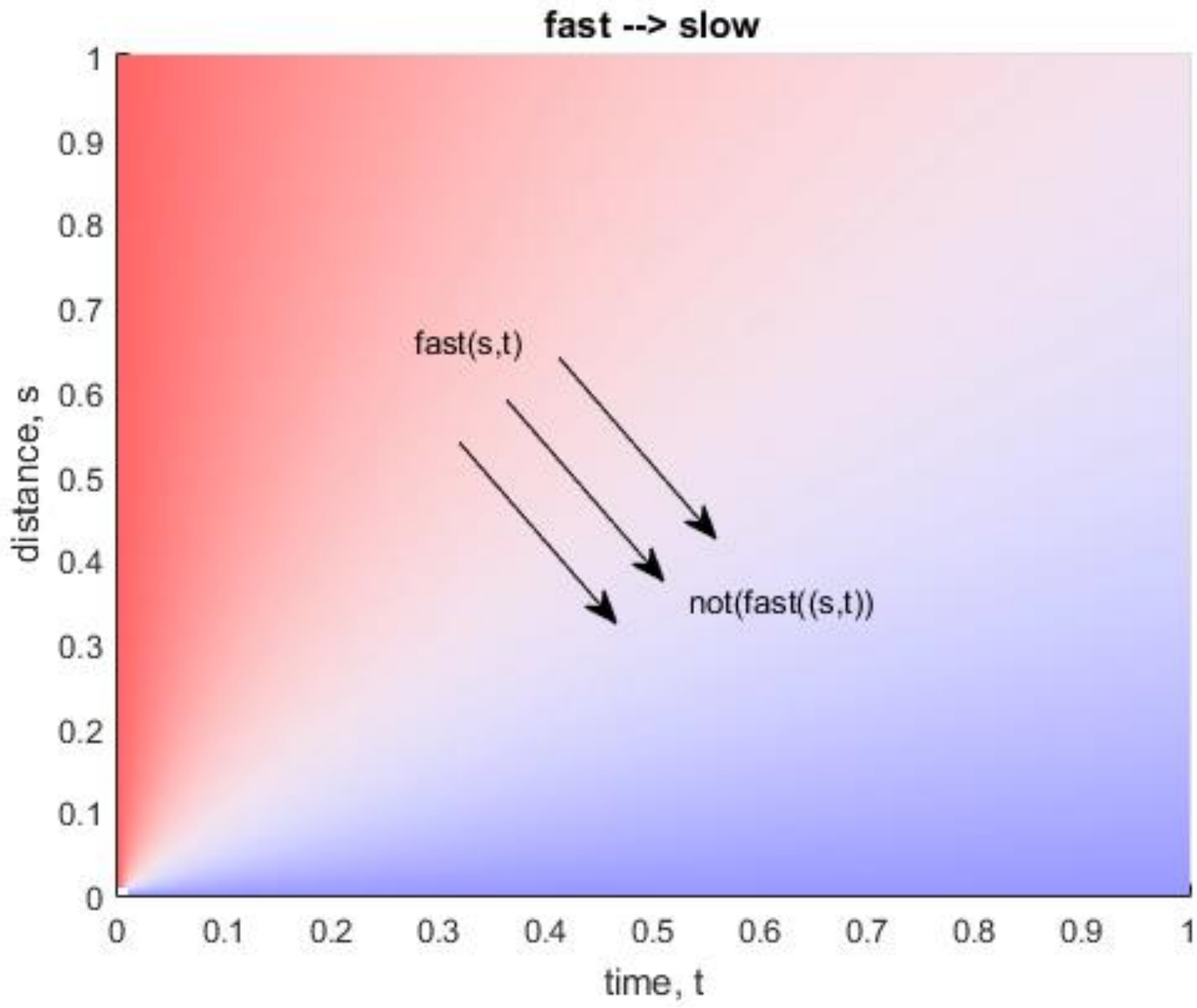

*Fig. 4-7. Mapping* slow = not(fast) *back to reference context*

We can define “very” approximately like this: $very(x) = x^2$ (Fig. 4-9). This is similar to Zadeh’s “very” [7:23].

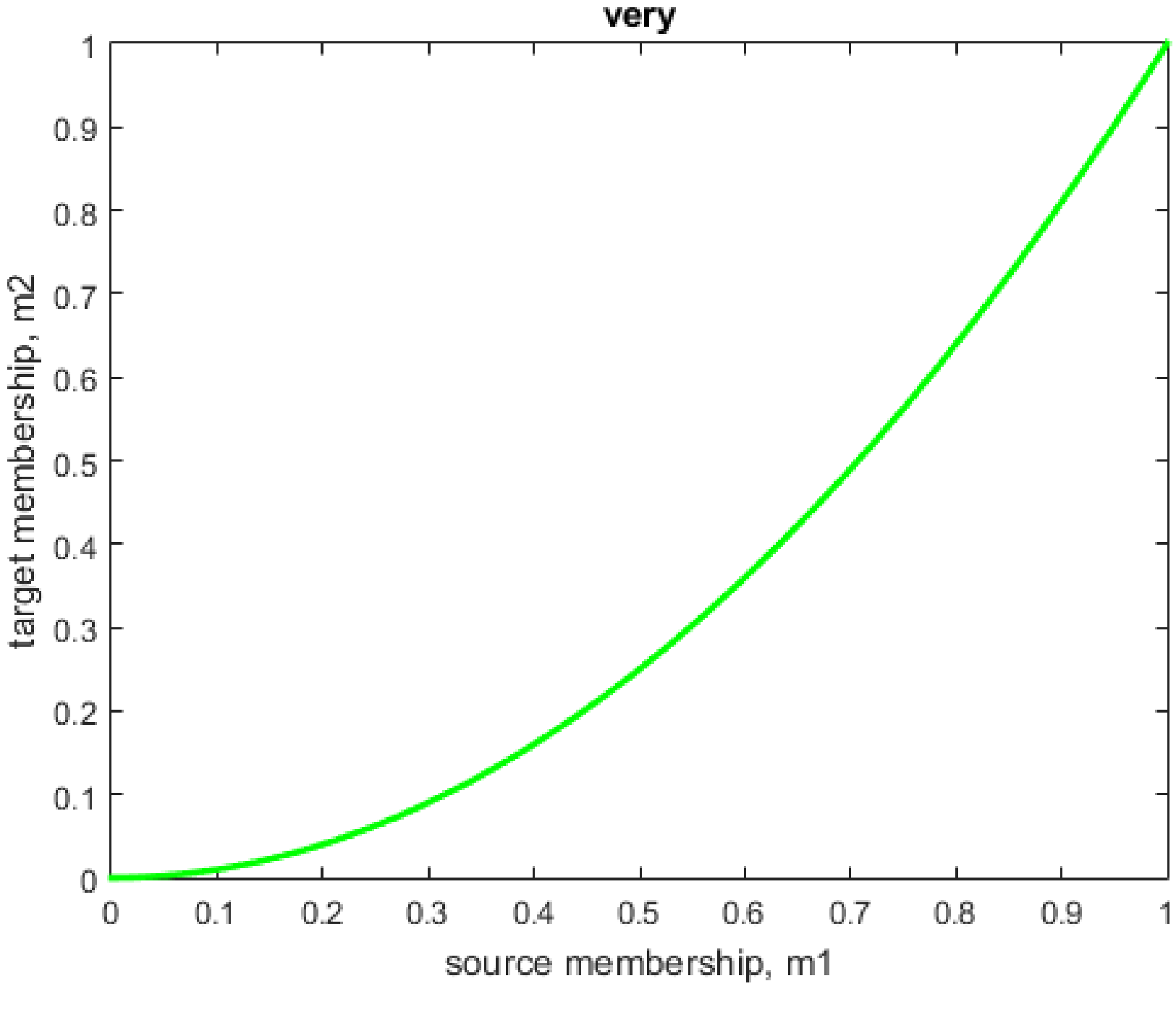

*Fig. 4-9. Operator "very"*

Applied to region "fast", we'll get region corresponding to "very fast" (Fig. 4-10).

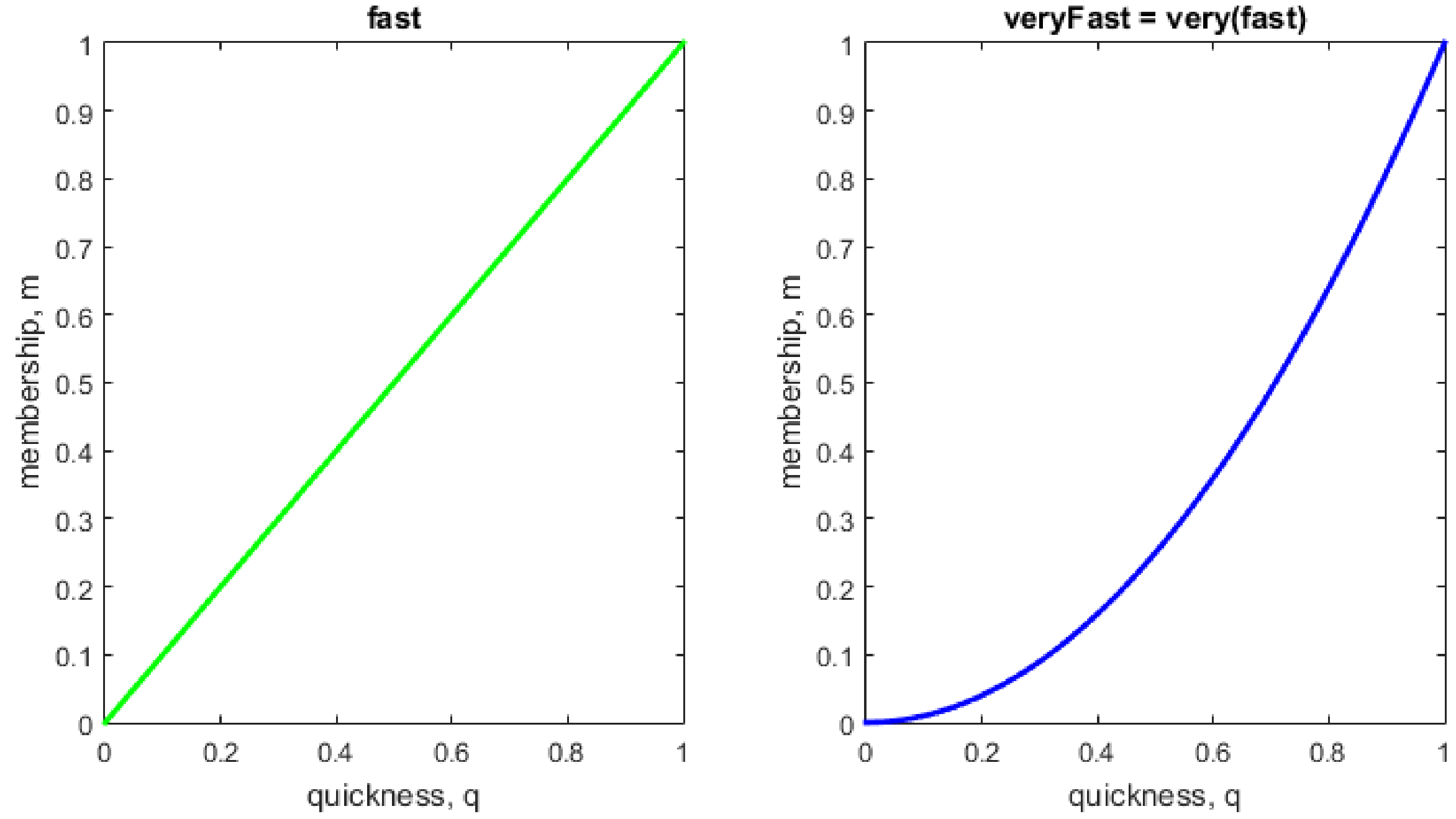

*Fig. 4-10. veryFast = very (fast).*

Mapping this back to our reference context, we get: $very(fast(s,t))$ (Fig. 4-11).

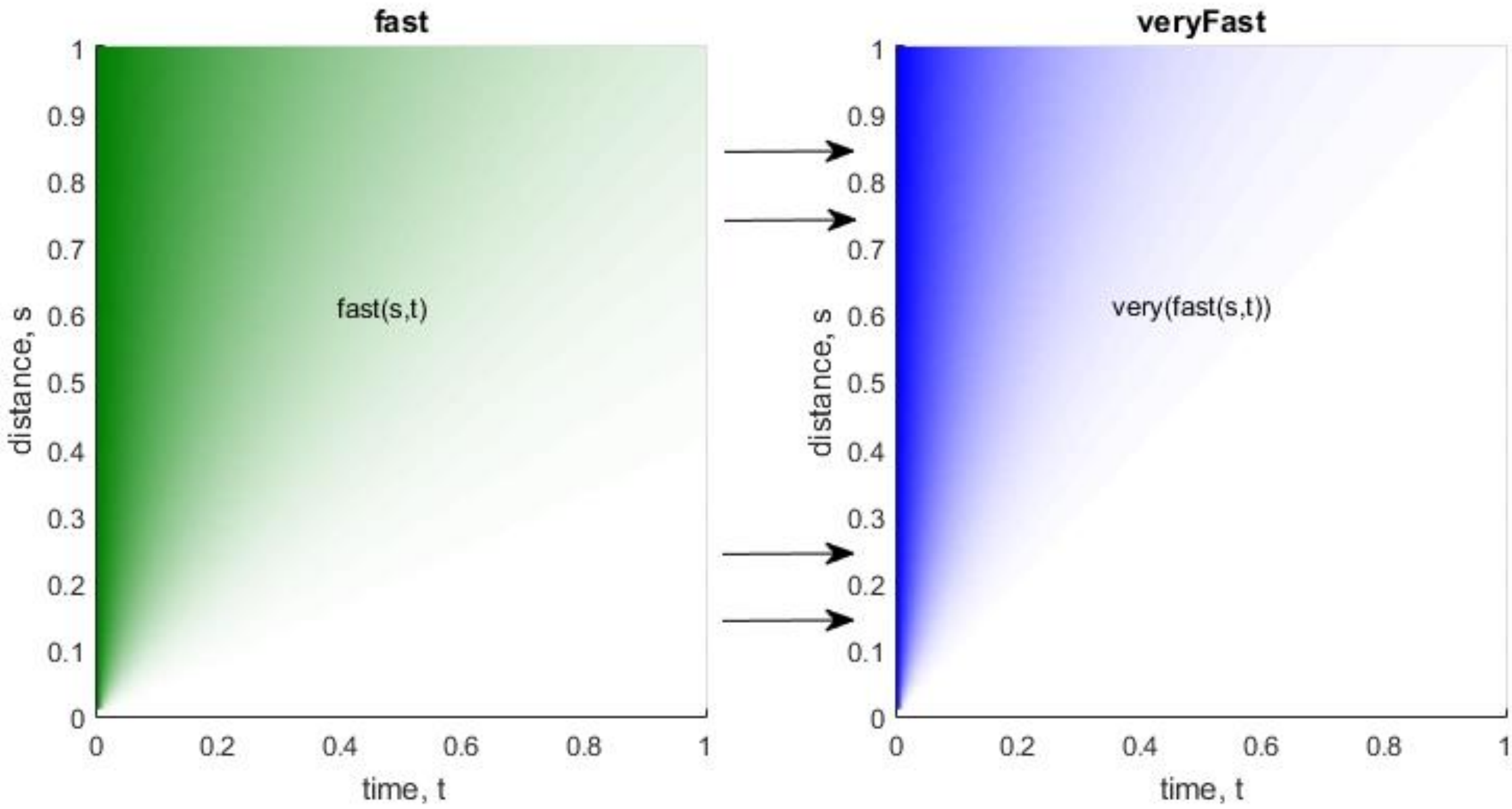

*Fig. 4-11. Mapping veryFast = very(fast) back to reference context*

# 5 Meaning operators

## 5.1 Parameter space

Let's now take a look at how we can use these ideas to work with meanings.

We assume that for our goals we can use systems having their state completely described with a finite number (N) of independent real parameters (properties)[3] ranging from zero to one. In other words – a point in an N-dimensional unit hypercube. In practice, as long as the state is never known exactly, we are going to deal with a fuzzy region in this cube instead of a point.

How much information do we need to encode these regions? Well, even one real parameter (by real, we mean a real number in mathematics, not a data type in programming) already carries infinite amount of information. However, we intend to manage with relatively small amount of information, encoding the meaning of a typical phrase. To ensure adequate performance on today's computers, we believe that it would be acceptable if this information measures some kilobytes, maybe tenths or hundreds of kilobytes, but hardly – megabytes.

In this case, we could estimate the number of allowed parameters, if we assume for simplicity that a typical region is simply connected and its typical shape is a polyhedron obtained from a parallelepiped by tilting its faces (we are never going to use this assumption later, except for this estimate). We need to store the position for each face and its tilt with respect to all the axes: this gives approximately 2N numbers per face. Total amount of faces is also 2N at max. This way, we are going to need around $4N^2$ real numbers to describe such a region. This means that if we are not willing to exceed 100 KB per region, we should not work with regions needing more than 30-50 parameters to describe.

---

[3] Here we should mention that neither choice of independent parameters, nor choice of basic parameters appropriate to describe any given fragment of the environment is trivial. Luckily, we can hopefully benefit from the other side of natural language complexity, and trust that the concepts suitable for deriving appropriate parameters and useful words-elements have already been worked out as a result of long running-in in natural language.

We assume that the system should understand the meaning of a phrase based on the available information about the environment around. Though, it is obvious that no simply shaped 30-dimensional region can describe a complex enough picture of the world. But a specific phrase should not change the whole picture of the world, either. On the contrary, we expect that each phrase should be allocated a subspace of relatively small dimensionality (say, 30 to 50), accommodating all of the modifications, while projection of the region onto the remaining dimensions will stay the same.

We are going to call the subspace containing region modifications *phrase context*. In practice, we are going to allocate the context with a certain excess; and if we pick a 30-dimensional subspace, but in reality the region only changes with respect to 10 dimensions, there is nothing wrong about it.

Let's note that normally we can store multi-dimensional regions as a set of independent regions in low-dimensional (one- and two-dimensional) spaces. This is described in more details in 5.5 and 5.6, including the mechanism for merging such regions into one multidimensional region. In the case when we need to project a multidimensional region onto a subspace of smaller dimensionality, we can just discard corresponding low-dimensional regions (that are not a part of the projection).

## 5.2 Meaning of phrases and words

So, *before* interpreting the phrase we have a certain projection of the system state onto a given context: let us call it *source region* $S_1$. *After* interpreting, the state will change, and so will its projection onto that context. Let us call this new projection *resulting region* $R_1$. Now, if the source state were $S_2$, the resulting state after interpretation would also be different, say, $R_2$. Thus, each set of possible source regions $\{S_i\}$ has a corresponding set of resulting regions $\{R_i\}$.

Now we are getting ready for a key definition: *phrase meaning in a given context* is an operator transforming every source region to a corresponding resulting region. In other words, phrase meaning is simply a mapping $\{S_i\} \to \{R_i\}$.

Obviously, we are not going to define such operators by specifying the sets of source and resulting regions. We will use more practical and descriptive ways, e.g. composition of several basic operators of different kinds.

So, we made an attempt to define the meaning of a phrase. But phrases are composed from words. How can we model the meaning of separate words, is there any conceptual difference between word and phrase meaning? Intuition is suggesting that it is different: many of the words cannot form complete phrases, and can only be used together with other words. We can formulate the most important distinction like this: generally speaking, words are operators that depend on some parameters. For example, let's take a command "walk". A robot receiving this command should start moving with some average speed. If it receives command "walk fast", the speed would be different: word "fast" modifies a parameter of the operator "walk". In case of the command "walk very fast", word "very" modifies a parameter of the operator "fast", that, in its turn, modifies the parameter of the operator "walk" (Fig. 5-1).

Now we should mention that a word can have many such parameters, and that they are, of course, fuzzy (as long as we are talking about natural language). In other words, we can say that a word operator has its own internal context[4], and the region, defined in this context, affects the operator. The operator resulting from such interaction of two or several modifying operators, we are going to call *block-operator*.

---

[4] It is important not to mix axes of this context (that can have separate, utility meaning) and axes of the contexts of the main message, in which the operators of complete phrases work.

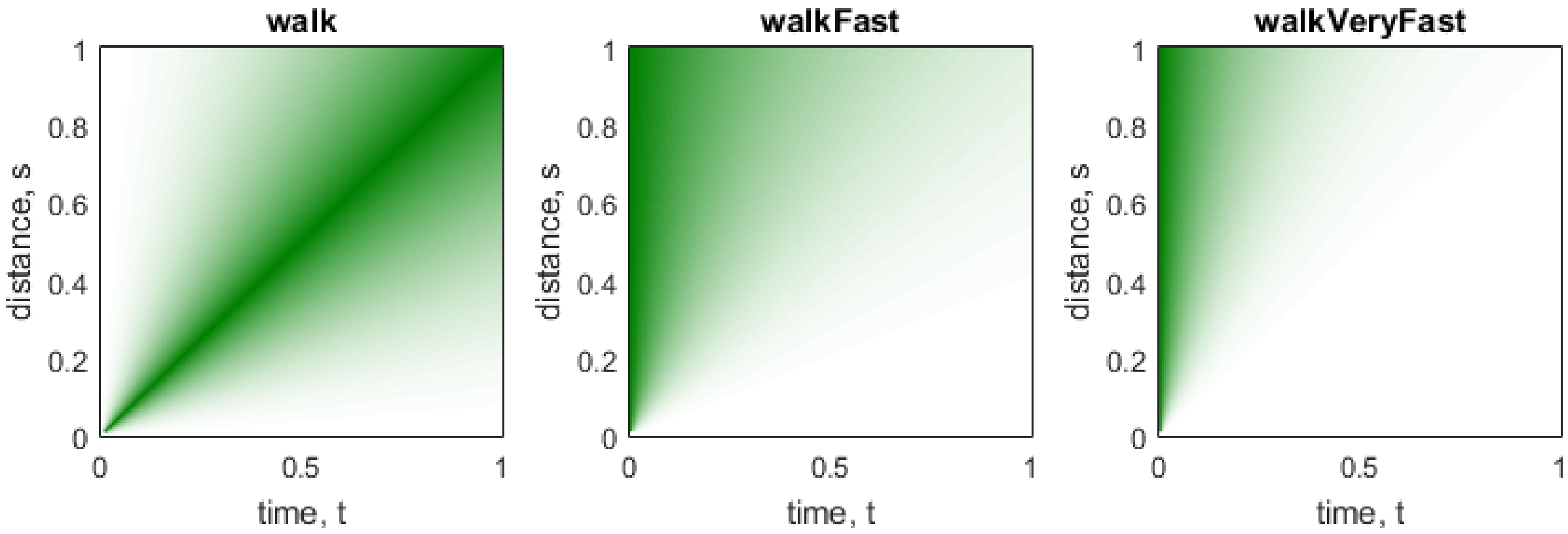

*Fig. 5-1. "Walk" -> "walk fast" -> "walk very fast"*

## 5.3 Different contexts

We have previously mentioned phrase contexts. However, much larger narratives (from a paragraph to a multi-volume novel) also have certain meaning. Variant reading is common when interpreting background, underplot, etc., but on the level "who went were, who did what and what were the consequences" even such large narratives can be fairly understandable. In our terminology this means that there is an operator transforming system state before interpreting the writing to the state after interpretation. Well, even if saying this is valid, in practice such representation is extremely redundant. In this case there will be hundreds of thousands of parameters characterizing the context, if not millions, and any practical work with operators in the space of such dimensionality becomes impossible.

In addition, as we mentioned before, the meaning is not obvious immediately, it is chosen for each individual phrase as result of comparing meaningfulness (comprehensibility) of the interpretations. And, if using brute force search through all phrase interpretations is perfectly possible, for large discourses it would be unfeasible.

The solution is that in case of independent statements (having to do with different objects, characters etc.) we can trust that the meaning-operators will be working in different subspaces. So, if we have two phrases *A* and *B*, their contexts are $C_A$and $C_B$, and their meanings are $A: X \to A(X)$ and $B: Y \to B(Y)$, then the meaning of the statement including both phrases is $A \oplus B: X \oplus Y \to A(X) \oplus B(Y)$. And even if we can work with each of these phrases in the "large" context, $C_A \oplus C_B$, in which their meanings are expressed with operators $A \oplus I: X \oplus Y \to A(X) \oplus Y$ and $I \oplus B: X \oplus Y \to X \oplus B(Y)$, it is obvious that it is just a waste of resources. Thus, in practice we are working separately with independent contexts, remembering that we can consider all transforms happening in one large space.

Now, when we have defined what we mean when talking about several contexts, let's discuss in more details how these contexts are brought into focus. They may be either found among the ones used before (i.e. we have already worked in this subspace), or created anew (i.e. we have not yet paid attention to this subspace of the "global context"). A typical situation can be described like this: first, we try to work in the same context we used for the previous phrase, but if we are running into understanding issues, we look for another suitable context, and if that fails – we create a new one. Ways of searching is a separate topic, for now let's just mention that the contexts can be organized in some kind of hierarchy for convenience. Moreover, we could create several hierarchical indexes: hierarchy of narrative parts, time intervals, spatial locations, objects, events, actions, etc.

## 5.4 Axis expansion

Let's assume that we have a property $A$ given by a reference region that is defined in paramers $x$ and y. Parameter $x$, in its turn, is defined by a region with coordinates $x1$ and $x2$, and parameter $y$, respectively – by a region with parameters $y1$ and $y2$. Is there an automatic way to express property $A$ in coordinates $x1$, $x2$, $y1$ and $y2$? It turns out that yes, and it is very simple. If region $A$ is expressed by the function $A(x,y)$, and regions $X$ and $Y$ – by functions $X(x1,x2)$ and $Y(y1,y2)$, respectively, then the expression sought for is $A(X(x1,x2),Y(y1,y2))$. Really, the value of membership function for the region $X$ shows the truth degree of parameter $x$, and this is exactly what needs to be provided as the first argument of function $A$. The same is also true for the truth degree of parameter $y$. In this way, we can express any regions via basic parameters. Some examples of axis expansion were given in 4.1 and 4.2.

An important comment should be made here: in this case certain basic parameters (e.g. related to time and space) may appear several times in the context, and with different scales (in the sence that the value on the time axis may in one case mean "hour", and in another case – year). This should not worry us too much: for example, color of different objects may just as well be represented by two different axes, even if they have similar meaning. Additional information on the hierarchy of scales, objects, etc. can be stored in earlier mentioned (5.3) hierarchical stores of events, objects and contexts. As we commented previously, these are a practical way to optimize work, while keeping the information (that is not used in the context at this very moment) for later.

## 5.5 Specifying regions

It may be convenient to store regions using reference points, that are used as a basis for interpolation. That is why it is also practical to define regions in the same way. The difficulty here is that, even though we are going to store multidimensional regions, it is only natural (from the user interface point of view) to define one-dimensional and two-dimensional regions. On the other hand, we can always define a multidimensional region as a sort of product of one-dimensional and two-dimensional regions. More precisely, our context would then be a direct sum of low-dimensional (one- and two-dimensional) contexts $A_1 \oplus A_2 \oplus \ldots \oplus A_n$, and our membership degree for region points would be the geometric mean of the point membership degrees in each of the low-dimensional regions. The choice of geometric mean is motivated by the following:

a) if in any of the combined regions we have zero ("clearly does not belong"), it means that the result should be zero as well,
b) only if all of the regions have one ("clearly belongs"), the total membership degree would be one,
c) surfaces with equal membership degree (e.g. equal to $x$) include the points, whose membership degree in all of the combined regions is also equal to $x$,
d) increase in the number of axes being combined does not result in radical decrease of membership function values in all places where its value is different from one.

Here, especially in the case with many axes, there may occur a need to avoid "equalization" of their relative weights. Really, it may happen that a region is defined in a space of 30-50 axes, but the most important axes are only one or two. In this case, instead of geometric mean we can take

$$\prod_{i=1}^{n} (x_i)^{\alpha_i}$$

where $x_i$ are point membership degrees in each of the combined regions, and $\alpha_i$ – a set of numbers (exponents) such that $\alpha_i > 0$ and $\sum \alpha_i = 1$. However, the last condition (normalization of the exponents) may be dropped – in this case, we would lose property c), but this may be convenient for definition of some regions.

It is worth noting that we do not always need to immediately combine low-dimensional regions, creating one region in a multidimensional context. As long as this is possible, it may be convenient to store the region as a set of several independent low-dimensional regions. This may be especially relevant when it comes to subsequent application of meaning-operators, for example qualitative adjectives (5.6).

Sometimes, we may need to define an unspecified (empty) region that corresponds to zero information. The values of the corresponding membership function will be equal to one for all the points of the region. It is easy to see that such membership function can be "added" to an existing region as a subspace, without changing existing properties of the region / its membership function.

Thus, we have a way of defining a region, and therefore, a projection operator (transforming any region into the given one). If we need to define a non-trivial operator describing region transformations, then, in addition to the source region, we can specify the trajectory of movement of reference points (e.g. using a spline), and the membership function degree change along the trajectory (e.g. using a different spline).

## 5.6 Parts of speech

The problem of selecting suitable and universal "building blocks" for constructing all possible kinds of meanings is an extremely complex one. However, each natural language is one solution to this problem. Let us look at some of the most basic "building blocks" normally used in natural language.

*Qualitative adjectives* ("large", "tall", "simple"). Operator, corresponding to a qualitative adjective, may be modeled as a projection operator. More concretely, $I(X) \oplus P(Y)$, where $I(X)$ is identity operator, $P(Y)$ is projection operator, and $X(x)$ and $Y(y)$ are the membership functions in corresponding region subspaces. $P(Y)$ substitutes $Y(y)$ for some particular membership $Y^*(y)$. Here we assume that the source region is either given as a set of independent membership functions $X$ and $Y$ (as it was mentioned in 5.5), or may be decomposed into a set of such functions. In the case when such decomposition is not possible with reasonable accuracy, we should treat this as an ambiguity within this interpretation (this is described in more details in 6.1). If the context has an axis directly corresponding to the adjective, then dimensionality of *y* will be one, otherwise – more than one.

We considered some examples of modeling qualitative adjectives in 4.1.

*Comparative adjectives* ("larger", "taller", "simpler") can be modeled as a direct sum of some general operator $G(Y)$ and identity operator: $I(X) \oplus G(Y)$, where $X(x)$ and $Y(y)$ are membership functions in corresponding region subspaces. Here, same as in the previous paragraph, we also assume decomposability of the source region into independent membership functions.

In the end of 5.5 we briefly described how such operators (corresponding to comparative adjectives) could be defined.

When it comes to nouns, they either create a new object in the object hierarchy or "actualize" an already existing one. If a new object is being created, then the axes specific for this word are added to its context, and the membership is formed with the spatial and other important parameters of the object. We would have obtained the same region if we first created a default (empty) context and then applied a number of adjectives, describing the properties of this noun.

*Verbs* are special, because they always work with the time axis, and often – with its small part. Therefore, if nouns creates or finds special object context, then verbs create or find action context.

Some *conjunctions* like "and" and "or" can be defined fairly simply: we can combine resulting regions in the same context analogous to the way we previously combined low-dimensional regions into a region, defined in a multidimensional space.

For example, for conjunction "and", instead of obvious multiplication, we can take the geometric mean of two functions: $h(x) = \sqrt{f(x)\, g(x)}$.

For "or", in its turn, we can take $h(x) = 1 - \sqrt{\big(1 - f(x)\big)\big(1 - g(x)\big)}$, where $f(x)$ and $g(x)$ are membership functions in the source contexts, and $h(x)$ – membership function in the resulting context. Such definition, as we already mentioned previously in 5.5, allows to avoid significant decrease in the membership function values, when membership degrees in the source contexts are lower, than one.

Simple illustrations to "and" and "or" are given on Fig. 6-1 and Fig. 6-2.

"Not", obviously, should work like $h(x) = 1 - f(x)$. This is similar to Zadeh's "complementation" [7:10]. An illustration to "not" was given on Fig. 4-5.

"But", on the other hand, prevents "context interpenetration". What we mean by this is that the second phrase of the two, connected with the conjunction, is interpreted like if the first one was not there. More formally, if the source region was $A$, and the first part of the phrase transformed it into $B$, then the meaning-operator of the second phrase part $F$ performs transformation $F(A)$ (and not $F(B)$). After that the results $B$ and $F(A)$ are stored separately in the context hierarchy, becoming different subspaces of some conceptual "global context", that contains all the information known to the system.

We gave some examples of how different parts of speech and concrete words can be modeled with operators of different kinds. Now let us take a look at some other problems: analyzing comprehensibility of the interpretations, and the problem of describing situations and concepts known to the system with words.

# 6 Interpreting phrases

## 6.1 Process

Let's now briefly discuss the phrase interpretation process.

When a new phrase comes in, we can do the following. First, as much as it is possible, the syntactic structure of the phrase is obtained from the grammatical forms of words. In practice, this means determining the order of application of operators and block-operators to one another. In case when this structure is ambiguous, we need to remember the possible options so we can later compare them based on their comprehensibility (6.3), and select the best option.

After, we need to do sequential application of operators to one another. More precisely, the operators that modify the internal context of another operator, do that (yielding a block-operator), while the operators working only with the external context of the narrative (e.g. phrase context), form a line, in the order of application. The result is phrase operator: composition of several block-operators. This operator is later applied to the narrative context (by sequential application of block-operators from the line).

After this step, we need to resolve the ambiguities of the structure. More concretely, we need to assess the meaningfulness of the resulting construction given the context of the narrative. If none of the options reach the threshold, we try using one of the "spare contexts" (see below). If that fails as well – we need to ask for clarifications. If several options are above threshold, we choose the best one, and save the rest (or some of them) as "spare contexts".

The number of "spare contexts" kept can be an adjustable parameter. When a phrase is unclear, and asking for clarifications is not an option, we can increase this parameter and reinterpret some of the recent phrases again. If we still fail to understand the phrase, we can increase both this parameter, and the number of recent phrases to reinterpret.

Our approach allows to handle situations when the syntactic structure of the sentence cannot be completely restored from word order and grammatical forms, and for restoring this structure people would need to resort to semantics. This situation may often occur when using a voice interface. In our approach, we can try different variants of the structure (there are not too many of these because of grammar and syntax rules), calculate the comprehension level (6.3), and choose the interpretation that makes most sense.

After the understanding of the phrase is achieved, we move on to the next phrase.

## 6.2 Polysemy and homonyms

Let's say some words about working with polysemy, and homonyms in particular. In case of essentially different word meanings, we can store several different operators corresponding to the same word. In case of homonyms (when words are used in very different contexts), the right meaning can be detected almost immediately, by comparing axes of word's internal context with axes of the current context.

In the case of more similar meanings the internal contexts of the words will be fairly similar. Then we will need to take all possible meaning-operators for the word and "test" them as part of the phrase operator, choosing the one resulting in the best satisfaction of comprehension criteria. It would be best if we normally store such similar meanings as one "fuzzier" operator, as opposed to storing separate meanings as separate operators, otherwise there is a risk of ending up with too many variants of phrase meaning (if the phrase contains several such words with involved polysemy).

## 6.3 Assessing comprehension

In this section we describe some techniques and ideas that can be used to determine whether a statement or a phrase makes sense, and to choose which of the possible interpretations makes the most of it.

We can start with trying to define several heuristics to help identify vague, contradictory and other incomprehensible phrases.

A region that has no points with high enough degree of membership (say, 0.95) can correspond to a contradiction. Indeed, this situation would mean there is no property combination in the context that definitely corresponds to our concept. For example, if we consider a hypothetical concept *and(slow, fast)*, we get a typical contradiction (Fig. 6-1).

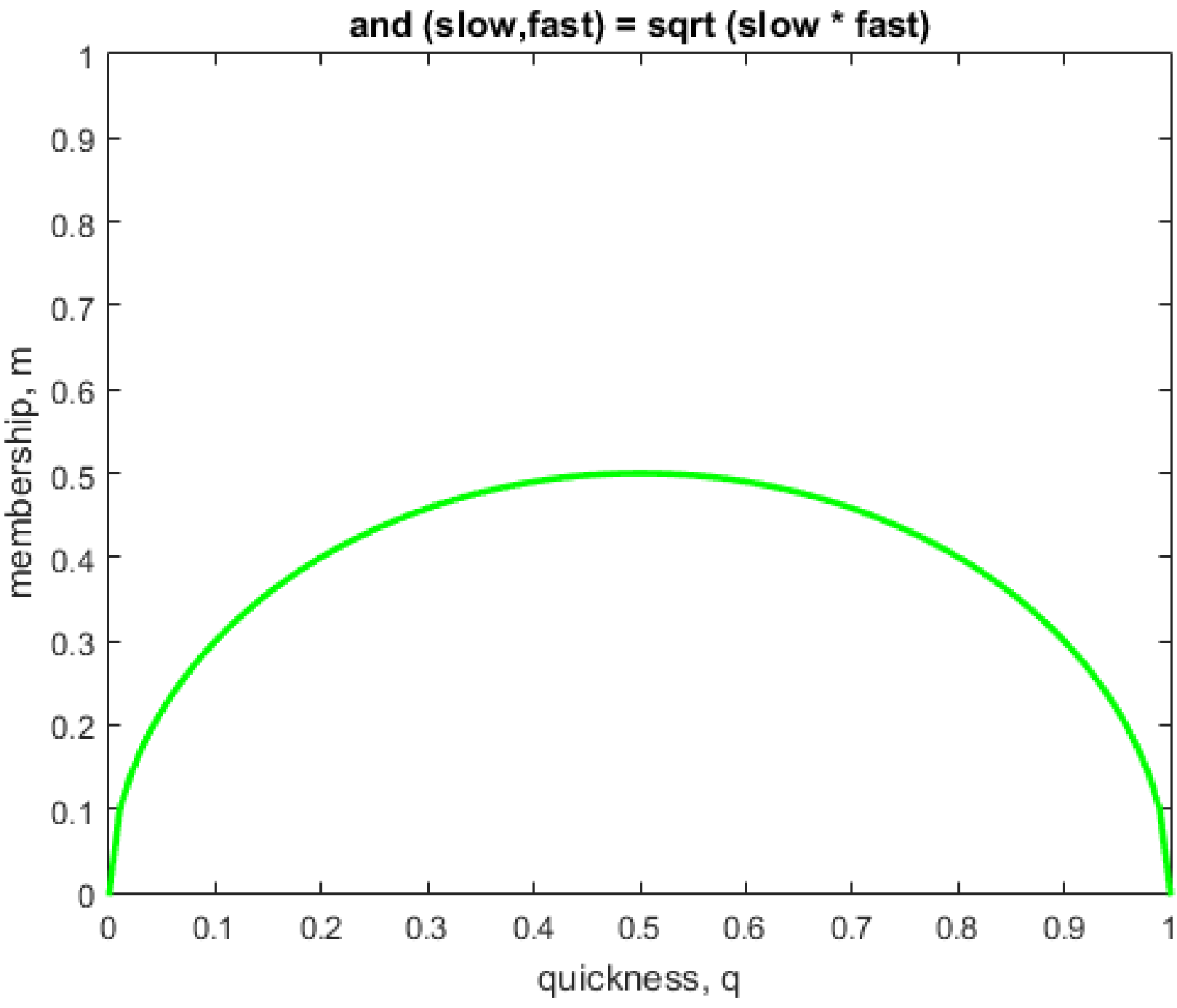

*Fig. 6-1. Contradiction*

On the other hand, if almost all the points of the context belong to a region, this usually does not make much sense either, as it provides no information. For example, if we consider hypothetical concept *or(slow, fast)*, we get into this situation (Fig. 6-2).

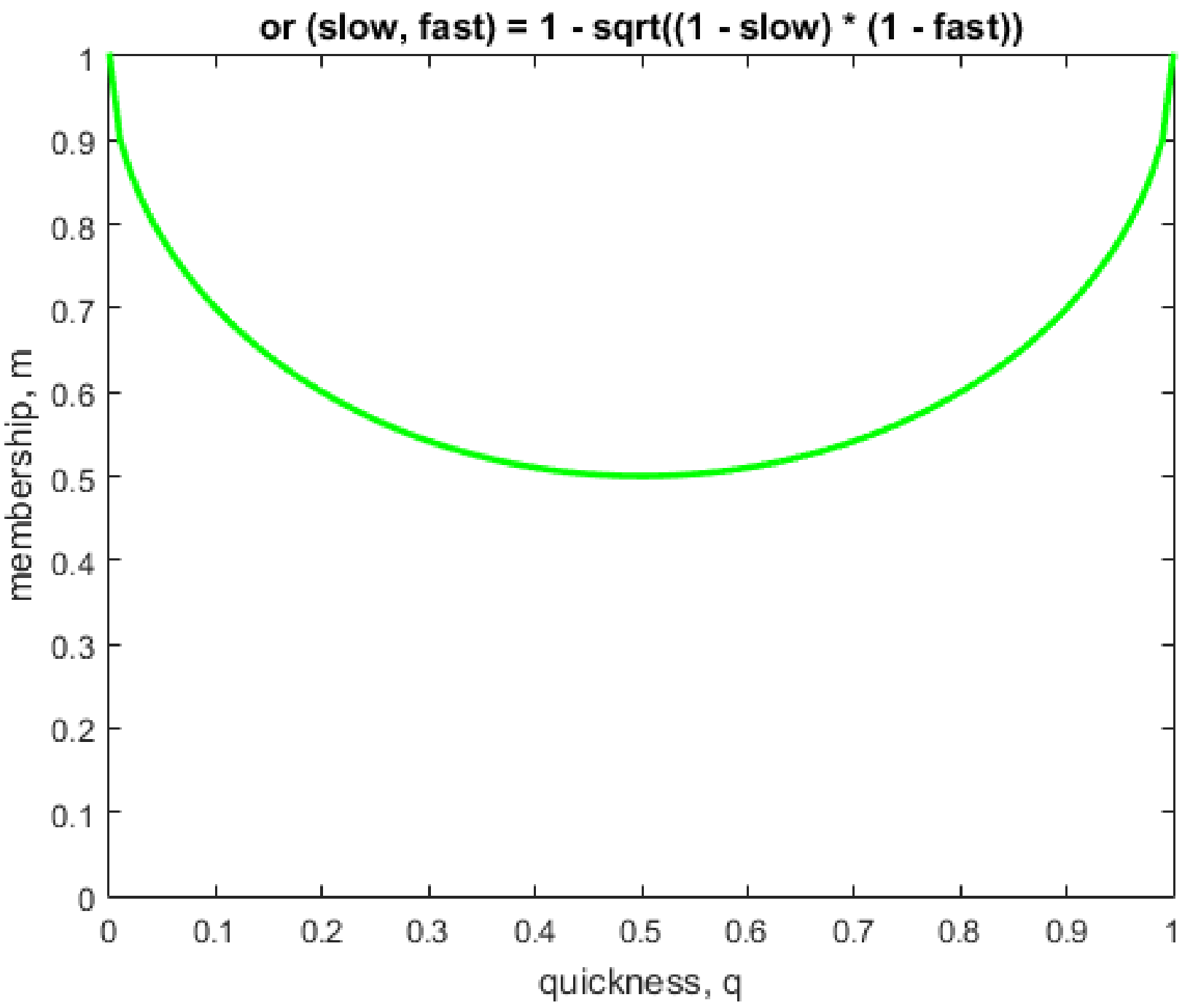

*Fig. 6-2. Lack of information*

Another useful heuristics has to do with the fact that normally we expect new information to change something in our knowledge. Let's compare the regions, describing verb-like concepts "walk" and "stand still" (Fig. 6-3).

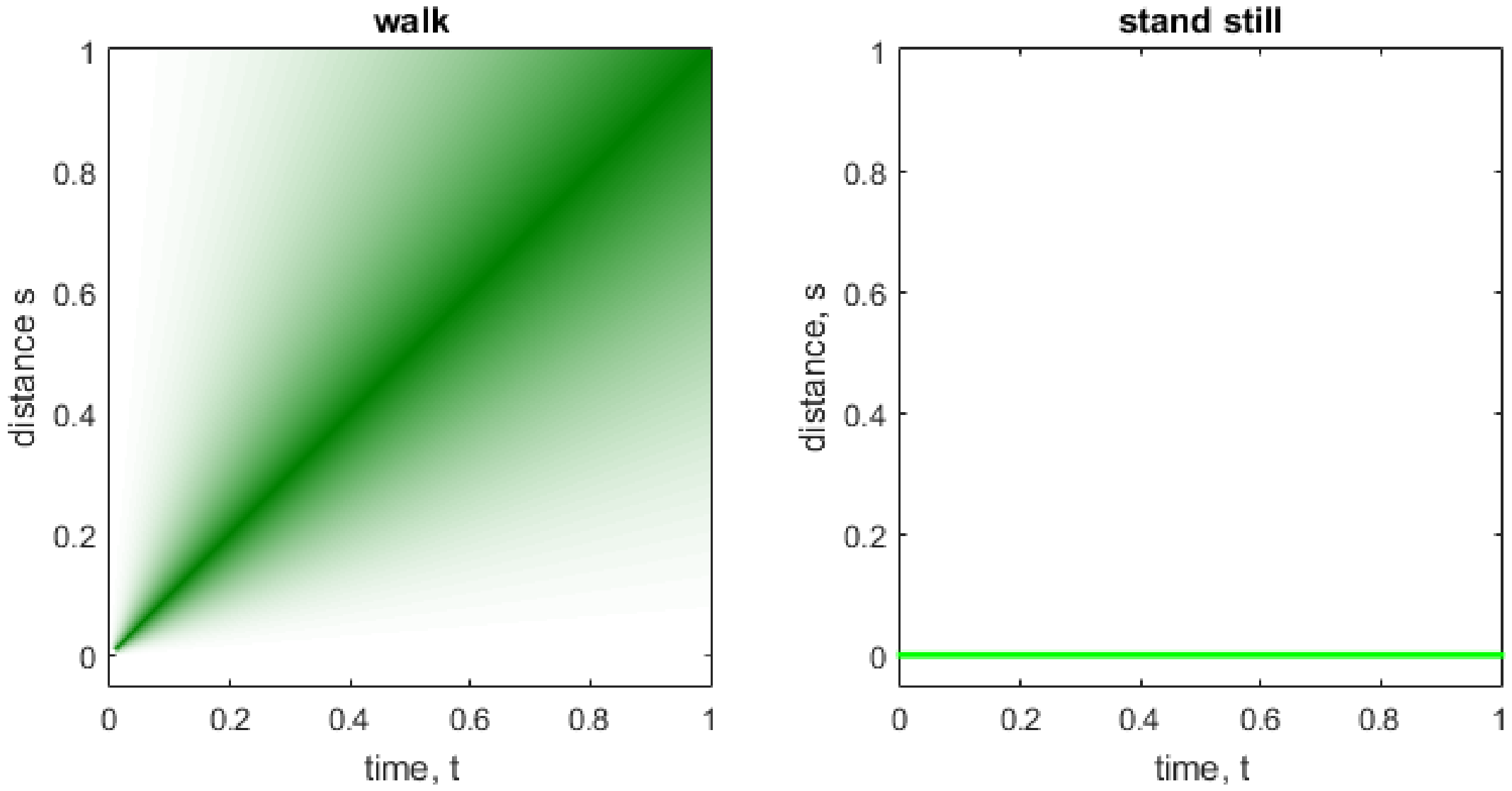

*Fig. 6-3. "Walk" and "stand still"*

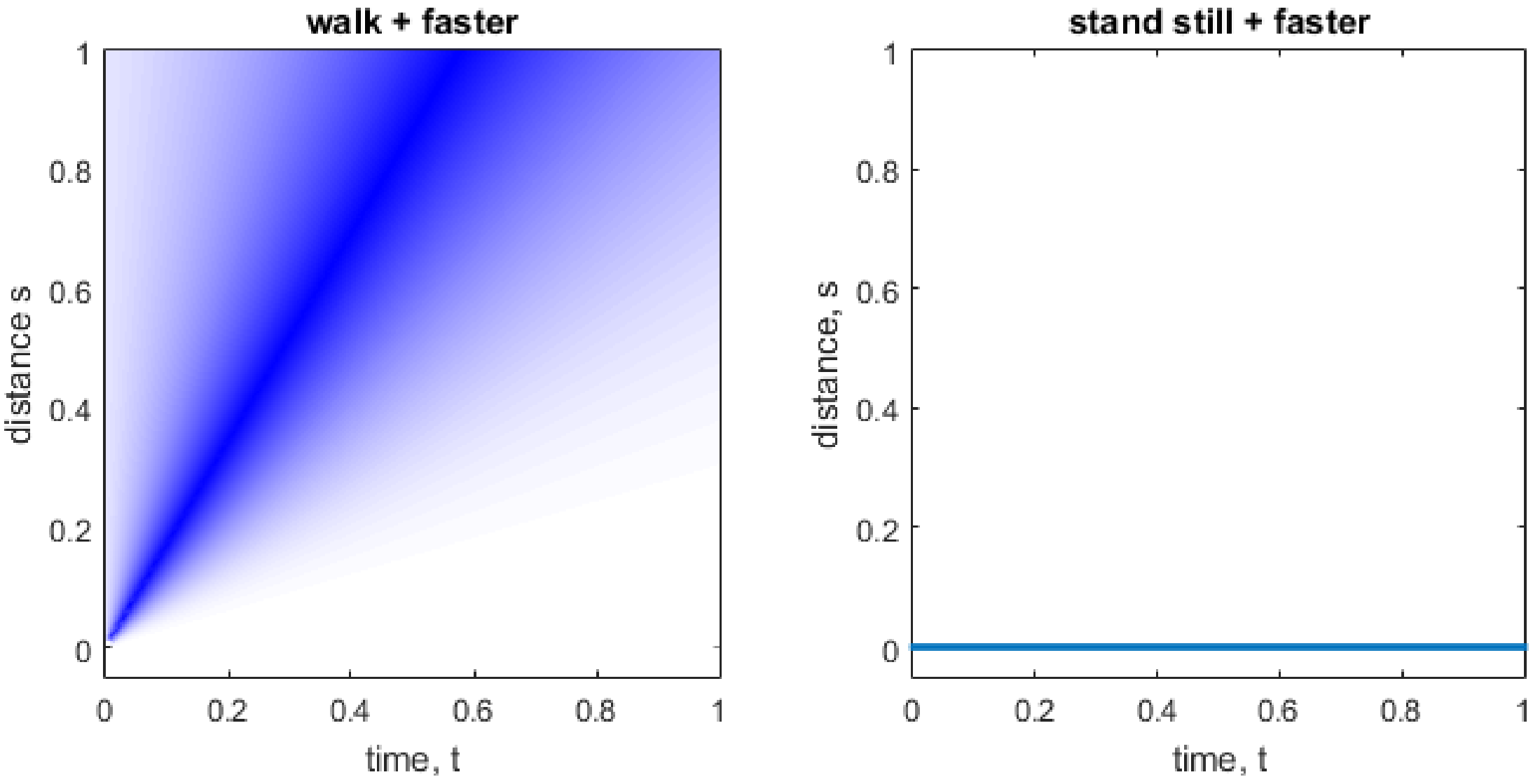

*Fig. 6-4. "Walk" + "faster" and "stand still" + "faster"*

If we apply operator "faster" to these concepts, "walk faster" will result in a different region, while "stand still faster" will remain the same, because the "compression" over time axis done by operator "faster" will have no effect (Fig. 6-4). So, after comparing the regions, we can conclude that phrase "stand still faster" makes no sense, indeed.

In many cases, new information not only changes our knowledge, but often is expected to precisiate it rather than making it more vague. This is especially the case when a system is receiving instructions.

For example, if we are trying to direct a robot, and the region describing it is transformed from "Somewhere NE" to "Anywhere except SW", it may be a sign of misunderstanding (Fig. 6-5). An exception to this could be a special phrase "forget everything, I will try to explain from the beginning".

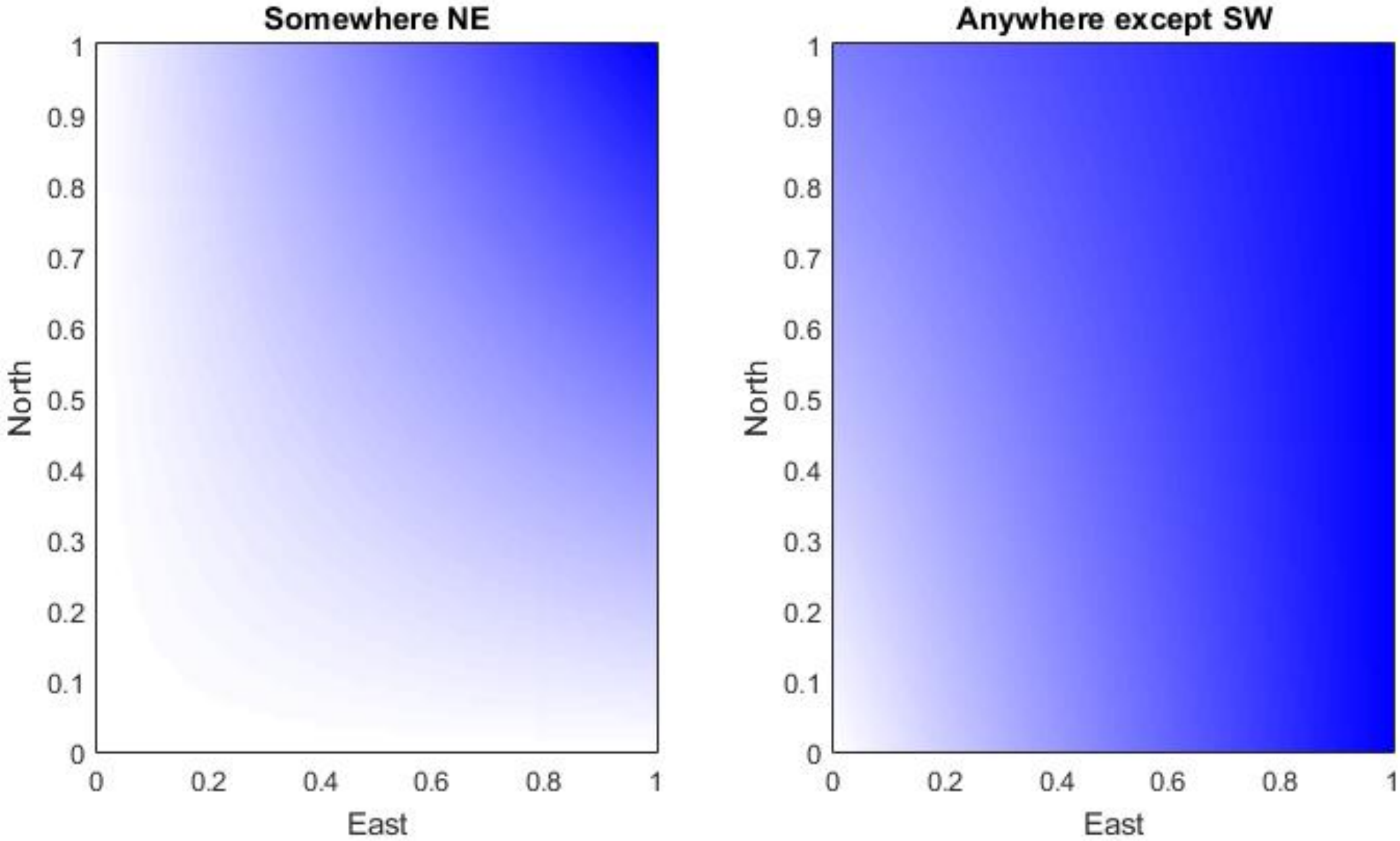

*Fig. 6-5. Description becoming more vague*

An important type of sensibility assessment is to analyze the correspondence between statement's assumed goal, and the information it actually conveys. At this stage of model development, we can, for example, check whether the grammatical mood of the message (like realis, imperative, conditional etc.) corresponds to the resulting region. Grammatical mood can be obtained from the grammatical form of the words or syntax (from parser).

When it comes to the region shape, we need to note that when giving natural language commands to the system, some parameters may be connected to system effectors. Like, say, parameter "movement speed" can be connected to a controller altering speed. Then, if the region with membership value above certain threshold is small and simply connected (say, region, corresponding to a command "drive fast"), we could find the center *C* of this region and output command "apply value *C*" to the controller's effector. However, if we give a command "drive very fast or very slowly", it would be natural for the system to ask for clarifications, even though the phrase itself is clear (unlike phrase "drive very fast AND very slowly"). On the other hand, when we are in realis or conditional mood ("but if I was driving very fast or very slowly"), such region shape is perfectly acceptable and does not require any clarifications.

It should be noted that many other, more specific heuristics can be created for estimating the comprehension level.

# 7 Composing descriptions

## 7.1 Abstract concepts

In 4.1 and 5.4 we considered how derived concepts (concepts defined via other concepts) are expanded, resulting in simpler concepts (defined via basic system parameters). Let's now take a look at how the system may derive new concepts, performing the operation of abstracting.

Let us consider two operations that we could perform with a region (or an operator). The first one is "blurring" that reduces boundary definition accuracy for the region (or image and preimage in case of operator). The other operation is dimensionality reduction: projection of this region onto a certain subspace (or restriction of an operator to this subspace). In both cases we reduce the amount of information used for defining the region (or operator).

It may happen that under "blurring" and/or dimensionality reduction, several regions (operators) become indistinguishable. And if we talk about an operator, the transformation defined by the operator can be correlated with a certain meaning. Then it may be possible to define a generalizing concept (that is described by the operator obtained with "blurring" or reducing dimensionality of the original operators), that will express certain common essence present in all of the considered operators.

More formally, let $X_i$ be a family of subspaces with a common subspace $Y$, $A_i$ – a family of operators in these spaces, $P_y: X \to Y$ – a projection operator onto subspace $Y$, and $\delta$ and $\varepsilon$ – accuracy levels with which we define parameter values and membership degree of region points.

Then, if an operator $B$ in a space $Y$ is satisfying

$$\forall i, x\ \exists\ \|\Delta y\| < \delta: \left\| B\left(P_y(x)\right) - P_y\big(A_i(x)\big) - \Delta y \right\| < \varepsilon,$$

we are going to call it *abstracting* with respect to meaning-operators $A_i$.

## 7.2 Describing with words

The problem of finding families of operators that have non-trivial abstracting meaning-operators is computationally quite complex. Despite being solvable in principle, building new meaningful abstractions in reality is going to be fairly difficult. However, here we can rely on natural language, hoping that all necessary abstractions are already available. Now we are going to talk about how these abstractions can be used in the problem of describing a given situation in words. Let's define the problem in more details.

We are going to distinguish three types of "describe in words" problem. First one – when there are many parameters known to the system, but unknown to the user. Possibly, the parameters are organized in a whole hierarchy of contexts, containing many hundreds and thousands of them. The problem is to convey this information to the user using minimal (or close to minimal) amount of words. Second type is similar to the first one, but in this case the information is not organized in contexts and parameters (in terms of our model). For example, it can be obtained from external sensors or be the result of solving some problem with a certain algorithm. In any case, it needs to be described in words. The third subtype of this problem is "interpretation crisis": none of the interpretations of what is said by the user does not meet the comprehension criteria. In this case, all of them should be described to the user, including the "failing" comprehension criterion (for each of them).

In all these cases the approach is similar: we need to obtain a meaning-operator transforming the source region (reflecting some knowledge taken as a starting point) into the resulting region that corresponds to the state being described. This correspondence may be partial or imprecise (correspondence of only a certain projection of the resulting region onto some subspace). In any case, we can define a measure of this correspondence, that is adequate for each concrete problem, using a special test operator that verifies the degree of goal satisfaction. And now we can define the problem of describing with words like the following: assume we have a certain original context $S$ with a defined in it region $A$ and a (large) set of meaning-operators $B_i$. These can be separate word operators, large block-operators, and even phrase and multi-phrase operators. Also, we have a test operator $G$ that verifies goal satisfaction: $G: F \to P$, transforming the region in the resulting context $F$ to the region in a one-dimensional context $P$ with the only parameter: goal satisfaction degree. The problem of describing in words is then composing (using the set $B_i$) such meaning-operator $D: S \to F$, that region $G(D(A))$ has values of membership degrees close to one when and only when the parameter "goal satisfaction degree" is also close to one (concrete values of proximity to one should be given separately, and may depend a lot on the problem in question).

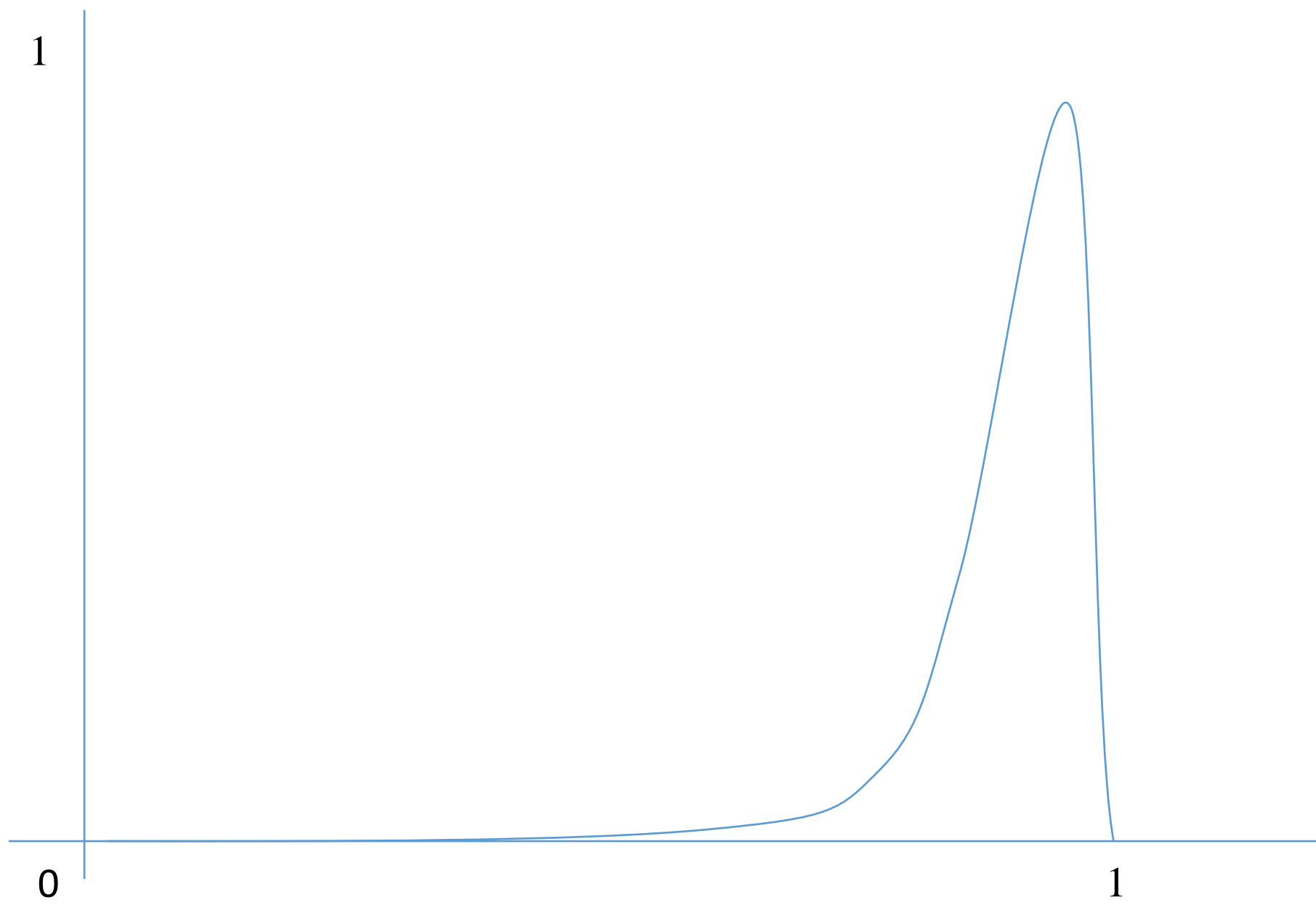

*Fig. 7-1. This is how we would like to see the membership function G(D(A))*

So, we have defined what we understand by "describing with words" problem. But, are there any ways to carry out this description, except for brute forcing all possible combinations of meaning-operators in $B_i$? We believe that yes, and that the hierarchy of abstract concepts already available in the natural languages may help us.

We saw that meaning-operators that are more abstract work in subspaces (contexts) of smaller dimensionality. That's why a suitable way of composing the description would be first taking the most abstract operators in the family $B_i$, making sure that at least some of them have parameters from the context $S$, and some – from the context *F.* There will be much fewer of such abstract operators, compared to the total number of operators. We can organize them in compositions $D_k = D_{k0} \circ D_{k1} \circ \ldots \circ D_{km}$, such that when adding a new operator to the composition, the region $G(D_k(A))$ would be shifting further towards one (Fig. 7-2). But we also expect, that when using the most abstract operators from $B_i$, the region $G(D_k(A))$ would be fairly fuzzy; this means that we cannot be sure that the goal satisfaction degree would close to one.

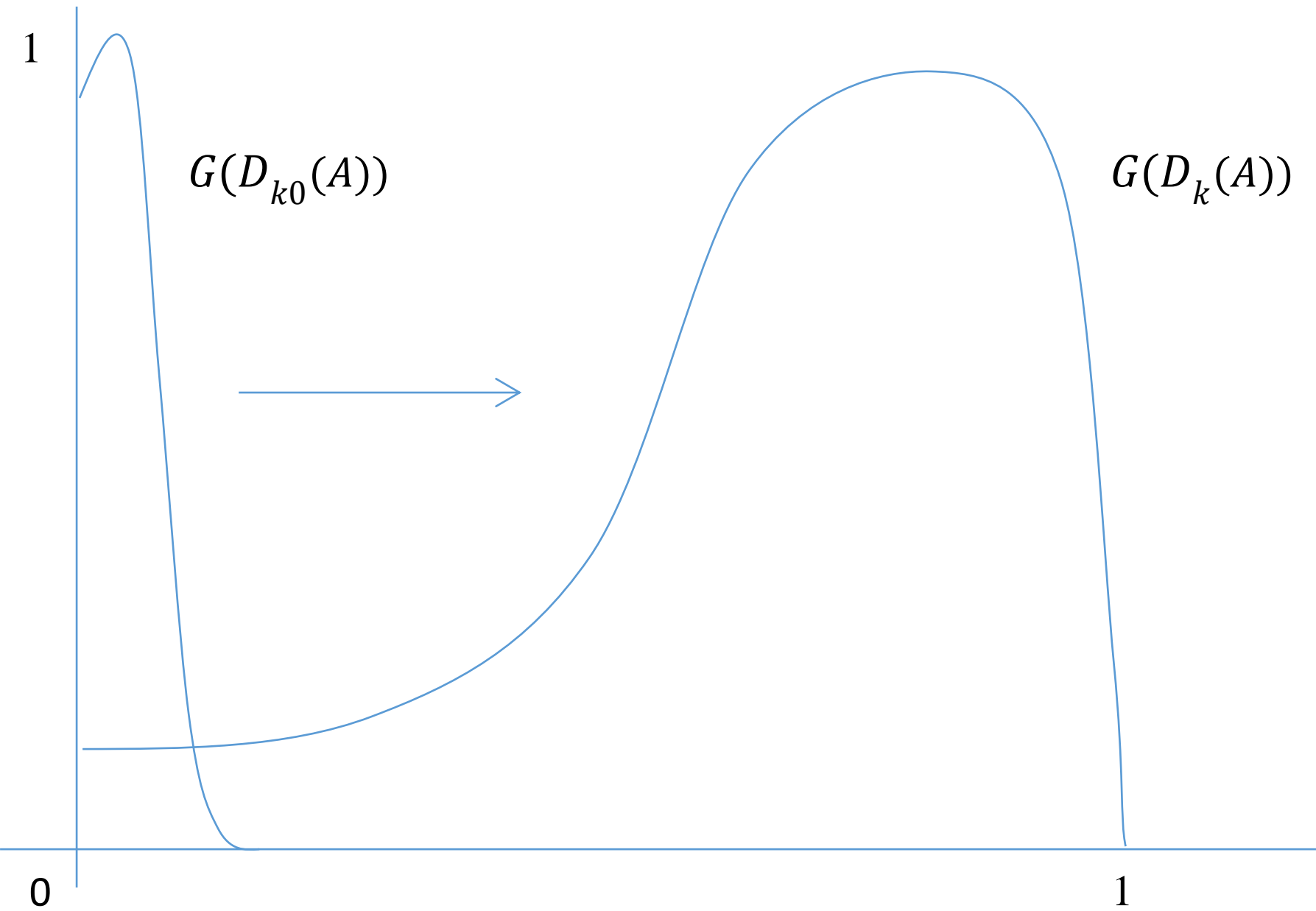

*Fig. 7-2. The change in the membership function of the "test region". Test region is a transformation (by the test operator) of the result of application of the first operator $D_{k0}$, and the whole composition $D_k$ to the region A. High abstraction level, high fuzziness.*

That's why we continue composing our description by making each of the composition components more specific. This means, that we move from composition $D_k = D_{k0} \circ D_{k1} \circ \ldots \circ D_{km}$ to the composition $D_l = D_{l0} \circ D_{l1} \circ \ldots \circ D_{ln}$, replacing each of the elements of the previous composition $D_{kj}$ with the sequence $D_{ls} \circ \ldots \circ D_{lt}$, such that operators $D_{ls}, \ldots, D_{lt}$ are less abstract, than $D_{kj}$ (have less fuzzier regions and operate in subspaces of larger dimensionality), while $D_{kj}$ is an abstracting operator with respect to the combined operator $D_{ls} \circ \ldots \circ D_{lt}$ (Fig. 7-3). Roughly speaking, operator $D_{ls} \circ \ldots \circ D_{lt}$ does the same as $D_{kj}$, but with some additional details. We expect that, as the abstraction level decreases, the test operator, applied to the region in the final context, would be yielding less fuzzier results, as shown on Fig. 7-4. If this is really the case, the process may be regarded as successful.

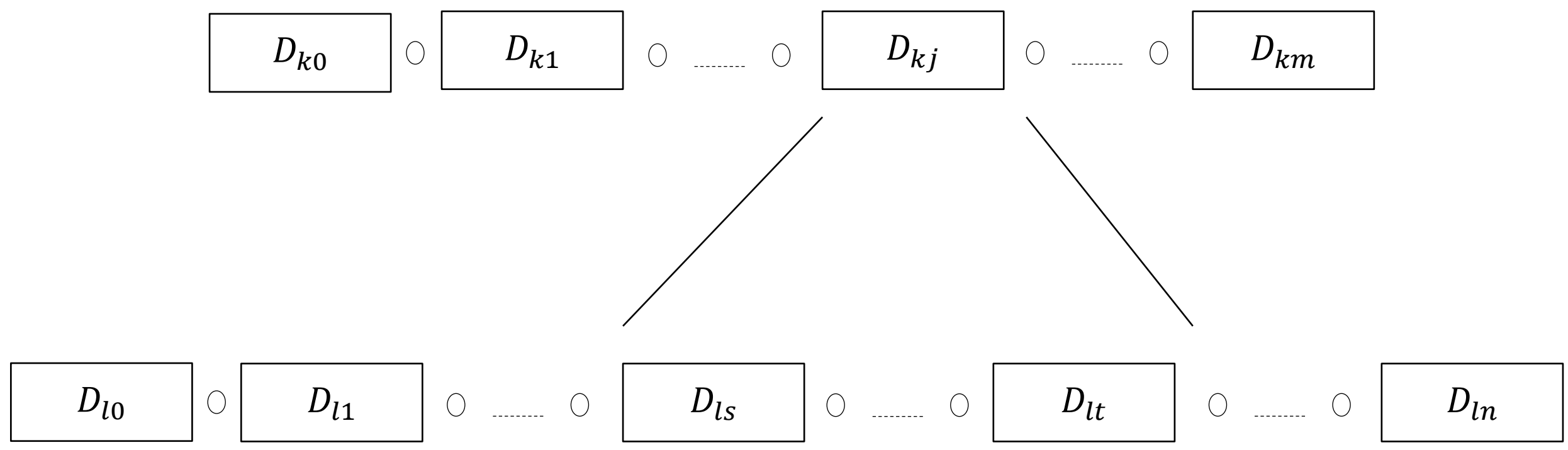

*Fig. 7-3. Substitution of a more abstract operator with a composition of less abstract operators.*

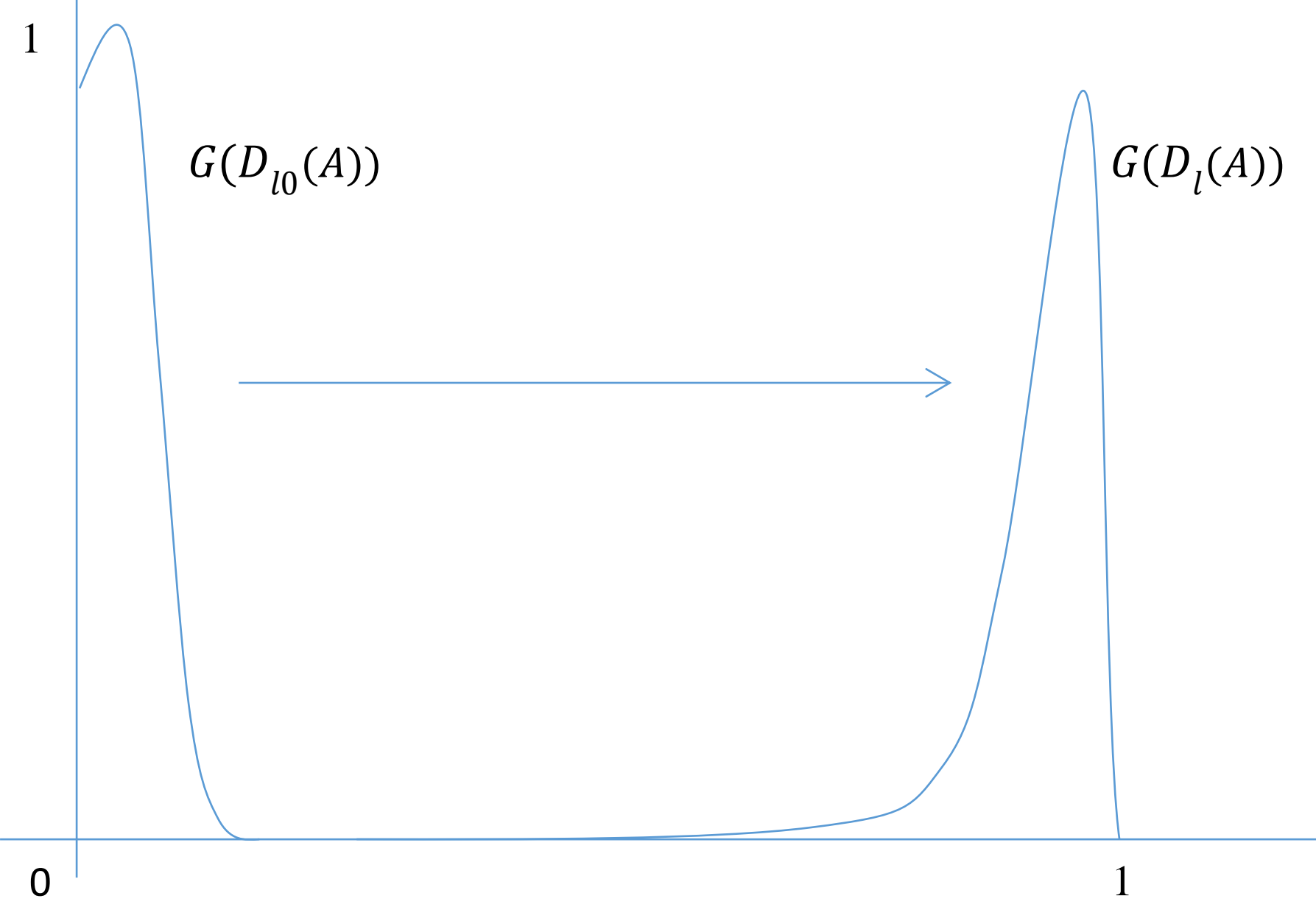

*Fig. 7-4. Expected behavior of the test operator for the first operator of the composition, and for the whole composition (sufficiently low abstraction level)*

It remains to discuss why we believe that such algorithm may lead to sufficient decrease in the total size of the enumeration. The idea is that at the highest level of abstraction, there are fairly few relevant operators (with parameters matching either source or resulting context, or – any other parameters in the operator composition so far). With each decrease in the abstraction level, we only talk about expressing (with less abstract operators) each of the more abstract operators in the composition so far – for what, again, we will only be attempting operators with sufficiently fitting parameter set. With each step of decreasing abstraction level, the fraction of "already defined" context parameters for the operators considered during the enumeration will be high – enough to decrease the enumeration to acceptable levels.

Thus, we demonstrated a draft of an algorithm for describing situations in words, based on the system of abstractions, available in the natural language. It is interesting to note that we can use this algorithm to describe the meaning of any operator or concept known to the system in "own words". For that, we need to exclude that operator (concept) from the set $B_i$, from which the meaning-operators used in the description algorithm are drawn. Such a description may turn out to be less precise, but possibly more compact (if it was originally given in words).

It remains to mention that when the system is experiencing difficulties with text comprehension, it may describe in words each of the interpretations, and more concretely: meaning of the largest narrative fragment, for which this interpretation succeeds the comprehension tests, and meaning of the smallest narrative fragment, for which the tests start failing. When the person communicating with the system is provided this data, she would have enough information to adjust the situation and rephrase her explanations or commands in a different, clearer way.

# 8 Conclusions

We described our approach to working with semantic information, that allows us to give definitions to word and phrase meaning, and formulate criteria for text comprehensibility. This possibility potentially allows our system to learn new words with help of texts meant for people (dictionaries, encyclopedias etc.), as long as certain minimal vocabulary is accumulated. Then, using the same types of sources, verification of the learning may be performed. For example, if a word just learned has a known synonym, we may verify closeness of the corresponding meaning-operators. In a similar way, meaning-operators obtained from the definitions given in different dictionaries may be verified.

A separate advantage of the method is the possibility to express a situation or a concept known to the system in words. It may allow for more productive dialog with a user, and her more complete awareness of the difficulties the system is experiencing.

Finally, storing often used block-operators, along with the ability to partially or fully "expand" them into more basic (or even built-in) operators, plus the abstracting procedure allows the system to obtain new and useful for it concepts. The procedure of "describing in words" may be applied to such concepts, followed by a "discussion" with the user, or a search in dictionaries. It may turn out that the system has discovered a concept that exists in natural language, or – come up with a new one. In principle, both possibilities are interesting.

# 9 Summary

The field of formalizing meaning and quantifying understanding is still much unexplored. In this introductory article we tried to make some humble steps in this field, describing our approach and ideas. Obviously, a lot of research and experimenting remains to be done before it is possible to evaluate any results. Having said that, we would like to emphasize that the approach described in this article has certain advantages, compared to those known and widely used. Semantic network-like representations can fairly easily absorb new knowledge, but the relations between words contain very little quantitative information. Other systems, like those based on different logic formalisms, encode information in details, but require human work for rule input and editing, and do not really handle uncertainty and imprecision well. Neither of the representations attempt to model the concepts for "what they are", and are not capturing their essence in quantitative form. This restricts capabilities of those approaches when it comes to complex understanding and reasoning related tasks (like engaging in a dialog). This is what our approach is focusing on, and we hope that its features will let it take its rightful place together with the other systems that obtain knowledge from natural language sources and allow communication with natural language.